\documentclass[aos,11pt,preprint]{imsart_arxiv}
\usepackage{amsmath}
\usepackage{amsfonts}
\usepackage{amssymb}
\usepackage{amsfonts}
\usepackage{amssymb}
\usepackage{amsthm}
\usepackage{mathrsfs}
\usepackage[pdftex]{graphicx}
\usepackage{enumerate}
\usepackage[numbers]{natbib}
\usepackage[colorlinks,citecolor=blue,urlcolor=blue]{hyperref}
\usepackage{caption}
\usepackage{subcaption}
\RequirePackage[OT1]{fontenc}
\allowdisplaybreaks

\startlocaldefs
\newcommand{\argmin}{\operatornamewithlimits{arg \,min}}
\newcommand{\argmax}{\operatornamewithlimits{arg \,max}}
\DeclareMathOperator{\tr}{tr}
\numberwithin{equation}{section}
\theoremstyle{plain}
\newtheorem{thm}{Theorem}
\newtheorem{lem}{Lemma}
\newtheorem{prop}{Proposition}

\begin{document}

\begin{frontmatter}

\title{Spectral and matrix factorization methods for consistent community detection in multi-layer networks}
\runtitle{spectral and matrix factorization methods}
\thankstext{T1}{Supported in part by NSF grant DMS-1406455.}




\author{\fnms{Subhadeep} \snm{Paul} \thanksref{m1}\ead[label=e1]{paul.963@osu.edu}}
\and
\author{\fnms{Yuguo} \snm{Chen} \thanksref{m2} \ead[label=e2]{yuguo@illinois.edu}}
\affiliation{The Ohio State University \thanksmark{m1} and University of Illinois at Urbana-Champaign \thanksmark{m2}}

\runauthor{S. Paul and Y. Chen}

\address{Subhadeep Paul\\Department of Statistics\\ The Ohio State University\\
Columbus, OH 43210\\ USA \\
\printead{e1}\\
}

\address{Yuguo Chen \\ Department of Statistics\\ University of Illinois at Urbana-Champaign\\
Champaign, IL 61820\\ USA \\
\printead{e2}\\
}

\begin{abstract}
We consider the problem of estimating a consensus community structure by combining information from multiple layers of a multi-layer network using methods based on the spectral clustering or a low-rank matrix factorization. As a general theme, these ``intermediate fusion" methods involve obtaining a low column rank matrix by optimizing an objective function and then using the columns of the matrix for clustering. However, the theoretical properties of these methods remain largely unexplored. In the absence of statistical guarantees on the objective functions, it is difficult to determine if the algorithms optimizing the objectives will return good community structures. We investigate the consistency properties of the global optimizer of some of these objective functions under the multi-layer stochastic blockmodel. For this purpose, we derive several new asymptotic results showing consistency of the intermediate fusion techniques along with the spectral clustering of mean adjacency matrix under a high dimensional setup, where the number of nodes, the number of layers and the number of communities of the multi-layer graph grow. Our numerical study shows that the intermediate fusion techniques outperform late fusion methods, namely spectral clustering on aggregate spectral kernel and module allegiance matrix in sparse networks, while they outperform the spectral clustering of mean adjacency matrix in multi-layer networks that contain layers with both homophilic and heterophilic communities.
\end{abstract}



\end{frontmatter}

\section{Introduction}

The study of multi-layer networks has received significant interest recently, driven by its myriad of applications in neuroscience, economics, genetics and social sciences \citep{mucha10,kivela14,boccaletti14,hxa14}. A multi-layer network is a powerful representation of relational data with the nodes representing the entities of interest and the network layers representing the multiple relations among those entities. While the term ``multi-layer network" is often used in a more general context, we focus our attention only on a network where the nodes are connected only within a layer and there are no inter-layer edges (such networks are also called ``multiplex networks" in the literature).

A dynamic or time-varying network represents different states of a single network over time. A dynamic network can also be represented as a multi-layer network, with the same node in consecutive time period usually being linked by an edge to respect the time ordering \citep{mucha10,bassett13,ghasemian15}. When appropriate for the application, e.g., in the problem of consensus community detection, we can ignore the time order and consider a time-varying network as a regular multi-layer network with no inter-layer edges \citep{hxa14,braun15}.

The problem of consensus community detection in multi-layer and dynamic networks has many important applications. Often in such networks one underlying community structure is in force while the different layers of interactions are merely different manifestations of the unobserved community structure.  For example, in the multi-layer twitter networks in \citet{gc13}, ground truth community memberships can be attributed to the users (nodes) based on attributes more fundamental and independent of the observed twitter interactions (e.g., political views, country of origin, football clubs), whereas the interactions provide multiple sources of information about the same latent community structure. Combining information from these multiple sources would then lead to enhanced performance in the learning task. Moreover, different representations of the same phenomenon often provide complimentary information, any one of which is not sufficient to describe the underlying process (see \citet{liu13} and the examples therein).

Even in situations where the hypothesis of a single constant community structure may not be true, e.g., in the analysis of dynamic brain networks, it is still often desirable to obtain a consensus partition that does not vary over time, but is a static average partition that remains in force throughout the experiment. Such an overall partition is crucial to obtain stable modules of brain regions as baseline for computing measures of local and global dynamism in the brain, e.g., ``flexiblity" and ``integration" in \citet{bassett11} and \citet{braun15}.

The present problem is also related to a more general class of problems that generally goes under the theme of multi-view clustering and has received considerable attention over the last decade, particularly in the computer science community. Numerous methods have been proposed to combine information from multiple views of a multi-view relational data for clustering. The goal is usually to leverage the diversity and often complimentary nature of the information in different layers to outperform simply summing the layers or using any one of the layers \citep{liu13}. A great many of those methods use spectral clustering or a low rank matrix factorization as a basis \citep{long06,zhou07,singh08,tang09,kumar2011co,ntk11,dong12,liu13}.

The ``linked matrix factorization" algorithm in \citet{tang09} and ``RESCAL" algorithm in \citet{ntk11}  approximate the adjacency matrices in each layer of a multi-layer graph, or each slice of a three way tensor, with a low rank symmetric matrix factorization. While one of the factors is shared the other one varies across layers or slices. Although the algorithms employed in the two papers are quite different, the factorization in both cases is computed by minimizing an identical joint Frobenius norm objective function. \citet{dong12} use similar common low rank matrix factorization ideas with a slightly different objective function to obtain a ``joint spectrum" of a multi-layer graph which is subsequently used for clustering.

The co-regularized spectral clustering in \citep{kumar2011co} with centroid based co-regularization maximizes the combined normalized cut objective function over the Laplacian matrices from all views of the data, subject to a smoothness penalty. This idea is similar to the evolutionary spectral clustering used in \citet{chi07} for clustering dynamic networks with a temporal smoothness penalty, and is part of a general theme of co-regularization in multi-view machine learning \citep{xu2013survey}. The co-regularization framework was extended to ``joint non-negative matrix factorization" using a Frobenius norm based objective function in \citet{liu13}. See \citet{sun13} and \citet{xu2013survey} for surveys of multi-view learning methods.

However, there is a lack of theoretical understanding of the objective functions in these spectral and matrix factorization based methods. Researchers often rely on simulations and applications to specific datasets to compare the methods. However, this approach fails to explore different scenarios that might arise in practice. For example, in multi-layer network applications, the component layers might have very different sparsity, signal quality and node degree distributions. Hence it is important to explore the utility of the methods under different statistical models and asymptotic settings through a principled theoretical study.

In this article we investigate the consistency properties of various methods for community detection under data generated from a multi-layer network model, the multi-layer stochastic blockmodel (MLSBM) \citep{valles14,hxa14,peixoto15,pc15,stanley15,pcmod,wilson2016community,barbillon2017stochastic}. We derive several asymptotic results to show consistency of the global optimizers of co-regularized spectral clustering and orthogonal linked matrix factorization under a high dimensional asymptotic setup where the number of nodes, the number of layers and the number of communities all grow. We use slight variations of the original algorithms to compute the solutions to the respective optimization problems. We note that both algorithms are not guaranteed to reach a global optimum. The present paper is an attempt to prove goodness of the objective functions rather than the algorithms, and is concerned with the following question: If it is possible to compute a global optimum or approximate one reasonably, will the global solution be consistent under a random graph model, namely the MLSBM?

In addition to the two methods mentioned above, we also consider two baseline methods previously used in literature. The first method is performing spectral clustering \citep{ng02spec,rcy11} on the mean of the adjacency matrices from different layers of the multi-layer network. This method has also been considered in \citet{tang09} and \citet{dong12} as a baseline method and is generally thought to be a simple but effective procedure \citep{kumar2011co}. In addition to including this method in our numerical comparisons, we also study its asymptotic consistency under MLSBM.

The second baseline method first computes a low dimensional spectral embedding (matrix of eigenvectors corresponding to top eigenvalues) and creates a spectral kernel for each layer, and then aggregates these kernel functions. A single layer community detection method is then applied to this aggregate spectral kernel \citep{tang09}. Another variation of this idea is to compute the community assignments in each layer independently using a single layer method (e.g., modularity or spectral clustering), and then create a ``module allegiance matrix" on which a single layer community detection method can be applied to compute the consensus communities \citep{braun15}.

The rest of the article is organized as follows. Section 2 describes the methods and algorithms considered in the article. Section 3 describes the MLSBM, defines mis-clustering rate and proves correct recovery in the noiseless case. Section 4 describes the consistency results. Section 5 contains a simulation study to numerically evaluate the methods. Section 6 gives concluding remarks.
All the proofs are given in the supplemental article \citep{supp}.

\section{Methods and algorithms}

We define an undirected multi-layer network with $M$ layers as a collection of graphs $\mathcal{G}=\{G^{(1)},\ldots, G^{(M)}\}$ over a common set of $n$ vertices. The vertices represent the entities/actors, while the layers represent different types of interactions among the entities. For the layer of the $m$th type, we define the adjacency matrix $A^{(m)}$ corresponding to that layer as follows:
$A_{ij}^{(m)}= 1$, if there is an edge of type $m$ between nodes $i$ and $j$, and $A_{ij}^{(m)}= 0$, otherwise.

We define the vector of degrees of node $i$ as $\mathbf{d_{i}}=\{d_{i}^{(m)} ; m=1, \ldots , M \} $, where $d_{i}^{(m)}=\sum_{j} A_{ij}^{(m)}$ is its degree of the $m$th type. Then the normalized graph Laplacian matrix for the $m$th layer can be defined as $ L^{(m)}=(D^{(m)})^{-1/2}A^{(m)}(D^{(m)})^{-1/2}$, where $D^{(m)}$ is a diagonal matrix with the degrees of the $m$th type of the nodes as elements, i.e., $D^{(m)}_{ii}=d^{(m)}_{i}$. Together the $M$ adjacency matrices create the three-way $n \times n \times M$ adjacency tensor of the multi-layer network $\mathbb{A}=\{A^{(1)},\ldots,A^{(M)}\}$. The corresponding Laplacian tensor is defined as $\mathbb{L}=\{L^{(1)},\ldots,L^{(M)}\}$. We denote the number of communities in the network by $k$. It will be assumed to be known throughout the paper. We use the notations $\|\cdot\|_2$, $\|\cdot\|_F$ and $\|\cdot\|_\Sigma$ to denote the spectral (operator) norm, the Frobenius norm and the trace norm, respectively, while $\tr(\cdot)$ denotes the matrix trace. We will use $\sin \Theta(U,V)$ to denote the diagonal matrix whose elements are sines of the principle angles between the subspaces $\mathcal{U}$ and  $\mathcal{V}$, spanned by the columns of the matrices $U$ and $V$ respectively (Definition 1.5.3 in \citet{stewart}).

We consider the following methods and algorithms for consensus clustering in multi-layer networks. The first two methods are so called ``intermediate fusion" techniques whereby the multiple layers are integrated through a clustering objective function \citep{liu13}. Such methods are often preferred over ``early" and ``late" fusion techniques due to superior performance \citep{vzitnik2015data}.

\subsection{ Linked matrix factorization}
 The first of the intermediate fusion methods is the linked matrix factorization (LMF) for clustering multiple graphs in \citet{tang09}. Our adaptation of the method is slightly different from the one described in \citep{tang09} in the sense that we enforce the columns of the shared factor to be strictly orthonormal and consequently drop the Frobenius norm regularization term (indeed this has been suggested in \citet{tang09}). In our simulations, we found the performance of both methods to be the same. To avoid confusion, we call our adaptation the orthogonal LMF (OLMF). Note that LMF has the identical objective function as the RESCAL algorithm, which is a three-way tensor factorization for learning in multi-relational data \citep{ntk11}. However the algorithm for RESCAL is different from that of LMF.

The OLMF solves the following optimization problem on the adjacency tensor of a multi-layer network:
\begin{equation}
    [\hat{P},(\hat{\Lambda}^{(1)},\ldots, \hat{\Lambda}^{(M)})] = \argmin_{P^TP=I} \sum_{m=1}^{M} \|A^{(m)} - P \Lambda^{(m)} P^{T} \|_F^2,
    \label{linked}
\end{equation}
where $P \in \mathbb{R}^{n \times k}$ is a common factor matrix and $\Lambda^{(m)} \in \mathbb{R}^{k \times k}$ are $M$ layer specific symmetric factor matrices.
This is equivalent to the following optimization problem (see the supplemental article \citep{supp} for a proof):
\begin{equation}
    \hat{P} = \argmax_{P^TP=I} \sum_{m=1}^{M} \|P^{T}A^{(m)}P\|_F^2, \quad \hat{\Lambda}^{(m)}= \hat{P}^{T}A^{(m)}\hat{P}, \ \ \, m=\{1,\ldots, M\}.
    \label{lmf}
\end{equation}

We will refer the objective function in (\ref{lmf}) as $F(\mathbb{A},P)$. While we require $P$ to have orthonormal columns, we do not put any constraint on the $\Lambda^{(m)}$ matrices, and specifically we do not require them to be diagonal matrices. Note that in general $[\hat{P},\hat{\Lambda}^{(m)}]$ is not the solution of the problem of finding the best at most rank $k$ approximating matrix for $L^{(m)}$. Hence in general, the matrices $\hat{\Lambda}^{(m)}$ are not the diagonal matrices of singular values. Intuitively the shared factor $P$ is expected to capture the common characteristics of the nodes in a multi-layer network including the latent community structure, while the different $\Lambda^{(m)}$ matrices capture the layer/relation specific characteristics.

We propose a BFGS algorithm to solve the OLMF optimization problem, similar to the algorithm in \citet{tang09}. The gradients are given by
\begin{align*}
    \frac{\partial O}{\partial P} & := -\sum_{m} (I-PP^T)A^{(m)}P\Lambda^{(m)}, \\
    \frac{\partial O}{\partial \Lambda^{(m)}} & := -P^T(A^{(m)}-P\Lambda^{(m)}P^T)P, \quad m=1,\ldots,M,
\end{align*}
where $O$ denotes the objective fuction in (\ref{linked}). Once the algorithm converges, we cluster the rows of the matrix $P$ using the k-means algorithm. Since each row in $P$ corresponds to one of the nodes, this gives a community assignment for the nodes.

\subsection{Co-regularized spectral clustering}
The second intermediate fusion method we study is the co-regularization based approach to multi-layer spectral clustering due to \citet{kumar2011co}. The idea of co-regularization has also been previously applied to various learning problems \citep{xu2013survey}. We adopt the centroid based co-regularization method from \citet{kumar2011co} unchanged in the context of multi-layer networks. The method, applied to the adjacency tensor, is based on solving the following optimization problem:
\begin{align}
  [\hat{U}^{(1)},\ldots, \hat{U}^{(M)},\hat{U}^{*}] = \argmax_{\substack{U^{(m)T}U^{(m)}=I, \, \forall m, \,\, \\ U^{*T}U^{*}=I}} \sum_{m=1}^{M} \{ & \tr(U^{(m)T}A^{(m)}U^{(m)}) \nonumber \\
  & + \gamma_{m}\tr(U^{*T}U^{(m)}U^{(m)T}U^{*})\},
  \label{coreg}
\end{align}
where $U^{(1)},\ldots,U^{(M)}$ and $U^{*}$ are $n \times k$ matrices with orthonormal columns. We denote $\mathbb{U}$ as the tensor containing the matrices $\{U^{(1)},\ldots,U^{(M)}\}$. The objective function of the optimization problem in (\ref{coreg}) is denoted as $F(\mathbb{A},\mathbb{U},U^{*})$. The optimization problem can be easily solved by alternating eigen decomposition of the matrices $A^{(m)}-\gamma_{m}U^{*}U^{*T}$ and $\sum_{m}\gamma_{m} U^{(m)}U^{(m)T}$ \citep{kumar2011co}. After the algorithm converges, consensus community assignments for the nodes can be obtained by clustering the rows of the matrix $\hat{U}^{*}$ with the k-means algorithm.

Note that the objective function contains two parts. The first part is the usual association cut spectral clustering objective function for different layers. The second part is a penalty function that seeks to maximize the cohesion between the eigenspaces obtained from different layers. To see this, we have the following proposition that characterizes the second part in terms of $\|\sin \Theta(U^{(m)},U^{*})\|_F$, which measures the distance between the column spaces spanned by $U^{(m)}$ and $U^{*}$ \citep{stewart}. The proof of this proposition, along with all lemmas and theorems, can be found in the supplemental article \citep{supp}.
\begin{prop}
For $U^{(m)}$ and $U^{*}$ as defined above, we have
\[
\tr(U^{*T}U^{(m)}U^{(m)T}U^{*})=k-\frac{1}{2}\|U^{*}U^{*T}- U^{(m)}U^{(m)T}\|_F^2=k-\|\sin \Theta(U^{(m)},U^{*})\|_F^2.
\]
\label{penalty}
\end{prop}

The penalty function alone is maximized when all the subspaces are identical, since $\|\sin \Theta(U^{(m)},U^{*})\|_F$ is $0$ when the subspaces spanned by $U^{(m)}$ and $U^{*}$ are identical \citep{stewart}. Hence the objective function represents a trade-off between optimizing the community structure in each layer (which might be noisy) and maintaining similarity with the mean community structure. The weights $\gamma_m$'s should be chosen to reflect both the desired trade-off between this two competing goals and the relative importance of the different layers. In particular, small values of $\gamma_m$'s will prevent sharing information across layers, which will result in estimates of $U^{(m)}$ being the one that is best for its own layer and the $U^{*}$ simply being the matrix of eigenvectors of $\sum_{m}\gamma_{m} U^{(m)}U^{(m)T}$. On the other hand, large values of $\gamma_m$'s will ensure the $U^{(m)}$'s try to achieve similarity with a common $U^{*}$ in expense of being sub-optimal for its own layer.



\subsection{Spectral clustering on mean adjacency matrix}

The first of the two baseline procedures we consider collapses the multi-layer network into a single layer network by taking the mean of the adjacency matrices from each of the layers. The usual single layer spectral clustering algorithm \citep{ng02spec,rcy11} is then applied to the resultant matrix. This procedure can be thought of as an ``early integration" or ``early fusion" technique, since data from multiple layers are aggregated before any processing is made \citep{vzitnik2015data}. Spectral clustering on some form of the aggregate matrix has appeared as a ``baseline procedure" in \citet{tang09,kumar2011co,dong12} and \citet{tang2012community}. In particular, consensus community detection proceeds through spectral clustering of the matrix $\bar{A} =\frac{1}{M}\sum_{m=1}^{M}A^{(m)}$. Consistency results for this method under the stochastic block model (SBM) were derived in \citet{hxa14} in the scenario when the number of layers grows but the number of nodes does not. \citet{chen2016multilayer} also derived phase transition results for a weighted version of this method under a model they characterize as ``multi-layer signal plus noise model".

\subsection{Aggregate spectral kernel and module allegiance matrix}
The other baseline method we consider is a ``late fusion" technique, where we first compute the eigenvector matrices $U^{(m)}$'s corresponding to the top $k$ eigenvalues from each of the $M$ layers of the graph and construct the aggregate spectral kernel matrix
\[
K_{n\times n} =\frac{1}{M}\sum_{m=1}^{M}U^{(m)}U^{(m)T}.
\]
However, instead of using kernel k-means to cluster the resulting matrix $K$ as in \citet{tang09} and \citet{dong12}, we apply spectral clustering to this matrix again to obtain the community assignments. We call this method ``aggregate spectral kernel". This is in spirit of clustering the ``module allegiance matrix" described in \citet{braun15}, where community assignment for each layer is first obtained using the Newman-Girvan modularity \citep{ng04}, and subsequently an $n \times n$ module allegiance matrix is formed, each of whose elements counts the number of times two nodes appear in the same module.

We use both the aggregate spectral kernel and the module allegiance matrix methods in our numerical study. It is worth pointing out that these methods are distinct from the majority voting method described in \citep{hxa14,pc15}. Although, much like the majority voting, these methods process each layer separately and fuse information later, one advantage is that both the aggregate spectral kernel and module allegiance matrix methods avoid the cumbersome issue of label switching ambiguity. To see this, assume we have two community assignment matrices $Z_1$ and $Z_2$ with $Z_1=Z_2P$, where $P$ is a permutation matrix, i.e., $Z_2$ gives the same community assignments as $Z_1$ but with its labels switched. However when we compute the module allegiance matrix, we have $Z_1Z_1^T=Z_2PP^TZ_2^T=Z_2Z_2^T$. The same is true for the aggregate spectral kernel. Intuitively, for each element they are concerned with whether two nodes belong to the same community or not, irrespective of which community that is. Hence they do not require solving a linear sum assignment problem as is required for majority voting.


\section{Models and mis-clustering}
The multi-layer stochastic block model (MLSBM) is a statistical model of multi-layer networks with a shared latent community structure \citep{hll83,hxa14,pc15}. We define the $k$ block, $M$ layer, $n$ node MLSBM as follows. Each node of the network is assigned a community label vector of length $k$, which takes the value of $1$ at the position corresponding to its community and $0$ in all other positions. Let $Z$ denote the $n \times k$ community assignment matrix whose $i$th row $Z_i$ is the community label vector for the $i$th node.

Given the community labels of two nodes, the edges between them in different layers are formed independently following a Bernoulli distribution with a probability that depends only on the community assignments and the relation the edge represents (i.e., type or layer of the edge). Hence within a community the nodes have ``stochastic equivalence" in the sense that the probability of an edge formation  (in any layer) with another node is the same for all the nodes in a community. We further assume that there is at least one node in each community which implies that there is at least one non-zero element in each column of $Z$.

The $k$ block, $M$ layer, $n$ node MLSBM with parameters $[Z,\mathbb{B}=\{B^{(1)},\ldots,B^{(M)}\}]$ can be written in the matrix form as
\begin{equation}
E(A^{(m)})=\mathcal{A}^{(m)}=ZB^{(m)}Z^{T}, \quad B^{(m)} \in [0,1]^{k \times k}, \, Z \in \{0,1\}^{n \times k},\label{SBM}
\end{equation}
where the matrices $B^{(m)}$ are $k\times k$ non-negative symmetric matrices of probabilities. For our analysis we will assume varying constraints on the rank of $B^{(m)}$'s. A similar rank based constraint is a standard assumption in the analysis of spectral clustering for single layer SBM as well \citep{rcy11,lei14}. We will refer to the matrix $\mathcal{A}^{(m)}$ as the population adjacency matrix for the $m$th layer and the tensor $\mathscr{A}=\{\mathcal{A}^{(1)},\ldots,\mathcal{A}^{(m)}\}$ as the population adjacency tensor.

\subsection{Correct recovery in the noiseless case}

Before we can tackle consistency of the methods, the first question that needs to be answered is whether a method can correctly recover the community assignments from the true population adjacency tensor when there is no sampling noise involved.
The following lemma shows that OLMF, co-regularized spectral clustering, spectral clustering of mean adjacency matrix, and aggregate spectral kernel, all can correctly identify the node community labels from the population adjacency tensor of MLSBM.

\begin{lem}
Let $\mathscr{A}=\{\mathcal{A}^{(1)},\ldots, \mathcal{A}^{(M)}\}$ be the three-way $n \times n \times M$ population adjacency tensor for MLSBM $[Z,\mathbb{B}]$ with each of the $M$ slices $\mathcal{A}^{(m)} \in \mathbb{R}^{n \times n}$ defined as in (\ref{SBM}). Then we have the following results:

(i) The optimization problem in (\ref{linked}) of orthogonal linked matrix factorization applied to the tensor $\mathscr{A}$ has $\bar{P}=ZQ^{-1/2},\bar{\Lambda}^{(m)}=Q^{1/2}B^{(m)}Q^{1/2},m=1,\ldots,M$, as the unique solution up to an orthogonal matrix, where $Q=Z^TZ$ provided at least one of the $B^{(m)}$'s is full rank. Further $Z_iQ^{-1/2}=Z_jQ^{-1/2}$ if and only if $Z_i=Z_j$.

(ii) The optimization problem in (\ref{coreg}) of co-regularized spectral clustering applied to the tensor $\mathscr{A}$ has $\bar{U}^{(m)}=Z\mu^{(m)}, \, m=1,\ldots,M, \, \bar{U}^{*}=ZQ$ as the unique solution up to an orthogonal matrix, where $\mu^{(m)}$ and $Q$ are invertible matrices provided each of the $B^{(m)}$'s is full rank. Further $Z_iQ=Z_jQ$ if and only if $Z_i=Z_j$.

(iii) The matrix containing the eigenvectors corresponding to the $k$ largest eigenvalues of $\bar{\mathcal{A}}=\frac{1}{M} \sum_{i=1}^{M} \mathcal{A}^{(m)}$ is $ZQ$ for some invertible matrix $Q \in \mathbb{R}^{k \times k}$ provided the matrix $\frac{1}{M} \sum_{i=1}^{M} B^{(m)}$ is full rank. Further $Z_iQ=Z_jQ$ if and only if $Z_i=Z_j$.

(iv) Define $\bar{K}=\frac{1}{M} \sum_{i=1}^{M} \bar{U}^{(m)}\bar{U}^{(m)T}$, where $\bar{U}^{(m)}$ is the matrix of eigenvectors corresponding to the largest $k$ eigenvalues of $\mathcal{A}^{(m)}$. The matrix containing the eigenvectors corresponding to the $k$ largest eigenvalues of  $\bar{K}$ is $ZQ$ for some invertible matrix $Q \in \mathbb{R}^{K \times K}$ provided each of the $B^{(m)}$'s is full rank. Further $Z_iQ=Z_jQ$ if and only if $Z_i=Z_j$.
\label{recoverySBM}
\end{lem}

We make two observations on the results of this lemma. First, note that in all of the above methods, the matrix whose rows are clustered using k-means algorithm for community detection has only $k$ distinct rows. Moreover, two rows are identical if and only if they are identical in the true community assignment matrix. This ensures that k-means algorithm in each case will correctly cluster the rows. Second, the methods require various conditions on the connectivity matrices $B^{(m)}$'s. In particular, the spectral clustering on mean adjacency matrix requires the aggregate connectivity matrix $\frac{1}{M} \sum_{i=1}^{M} B^{(m)}$ to be full rank, which is also related to the general issues associated with aggregating a multi-layer graph with diverse layers, explored from an information theoretical point in \citep{pc15}. Third, the noiseless recovery in co-regularized spectral clustering does not depend on what we choose for $\gamma_{m}$'s. This quite counter-intuitive phenomenon is true because both parts of the objective function, the association cut and the penalty term, are separately maximized by the true communities, with the penalty term achieving its global maximum irrespective of $\gamma_{m}$.

\subsection{Characterizing mis-clustering}
 Although Lemma \ref{recoverySBM} shows that the methods under consideration can perfectly recover community labels from the true population adjacency tensor, in reality we do not observe the true population tensor. Instead we observe a noisy sample version of it.  Consequently, community assignment using the methods will lead to some error. For a given benchmark community assignment and an estimated community assignment, we define a mis-clustering rate as the proportion of nodes for which the assignments do not agree. Let $\bar{e}$ denote the vector of true community labels extracted from $Z$ and $\hat{e}$ denote the vector of a candidate assignment. Then we define the mis-clustering rate
 \[
 r = \frac{1}{n} \inf_{\Pi} d_H(\bar{e},\Pi(\hat{e})),
 \]
where  $\Pi(\cdot)$ is a permutation of the labels, $d_H(\cdot,\cdot)$ is the Hamming distance between two vectors, and $\inf$ denotes the infimum over all permutations in $\Pi$.

Note that in each of the methods we consider, we obtain a low rank matrix with orthonormal columns whose rows are then clustered using the k-means algorithm for community detection. Hence we also need to relate this mis-clustering rate with the low rank matrices obtained from the methods. For a method under consideration, let $\hat{U}_{n \times k}$ be the low rank matrix with orthonormal columns it outputs, whose rows can subsequently be clustered to estimate community assignment $\hat{e}$. Then we have the following relationship,
\begin{equation}
r \leq \frac{8n_{\max}}{n}\|\hat{U}-Z(Z^TZ)^{-1/2}O\|_F^2,
\label{misclus}
\end{equation}
where $O$ is an arbitrary orthogonal matrix and $n_{\max}$ is the number of nodes in the largest true community \citep{rcy11}.

\section{Consistency results}

In this section  we investigate the asymptotic consistency of consensus community detection using the methods outlined in  Section 2. The asymptotic setup we consider is as follows. We let both $n$ and $M$ grow, and assume no relationship between their growth rate. However we will be most interested in the case when $M$ grows faster than $n$. This framework is particularly suitable for consensus community detection in dynamic graphs, where the number of layers represents the number of temporal snapshots available to us and can potentially be exponentially larger compared to the number of nodes. We also let the number of communities $k$ (which is assumed to be known in advance) to grow with both $n$ and $M$.

Before proceeding with the main results we prove the following theorem with two results on a multi-layer graph with independent edges, the first of which extends the results contained in \citet{chung11} to multi-layer graph settings and the second one is a new result using matrix Hoeffding bound \citep{tropp12}.
\begin{thm}
Let $\mathcal{G}$ be a multi-layer graph with each edge being independent of all other edges of all types. Let $\mathbb{A}=\{A^{(1)},\ldots,A^{(M)}\}$ be its adjacency tensor and let $\mathscr{A}=\{\mathcal{A}^{1},\ldots,\mathcal{A}^{(M)}\}$ denote the expected adjacency tensor. Further, let $\Delta_m$ be the maximum expected degree for a node in layer $m$. Define $\bar{\Delta}=\frac{1}{M}\sum_{m=1}^{M}\Delta_m$ and $\bar{\Delta}'=\frac{1}{M}\sum_{m=1}^{M}\Delta_m^2$. Then we have the following results:

(i) For any $\epsilon>0$, if $M\bar{\Delta}> \frac{4}{9} \log (2n/\epsilon)$, then with probability at least $1-\epsilon$,
\[
\|\frac{1}{M}\sum_m (A^{(m)}-\mathcal{A}^{(m)})\|_2 \leq  \sqrt{\frac{4\bar{\Delta}\log (2n/ \epsilon)}{M}}.
\]

(ii) If $
\frac{1}{M}\sum_m \exp \left(-\frac{4\Delta_m\log (2Mn^3)}{2\Delta_m+2\sqrt{4\Delta_m\log (2Mn^3)}/3}\right) \leq \exp(-\log (2Mn^3)) $, then with probability at least $1-o(1)$ we have
\[
\|\frac{1}{M}\sum_{m} (A^{(m)}-\mathcal{A}^{(m)})^2\|_2 \leq (\log n)^{(3+\delta)/2} \frac{\log 2M}{\sqrt{M}} \sqrt{\bar{\Delta}'} + \bar{\Delta}
\]
for some $\delta>0$.
\label{thm:lavrg}
\end{thm}

We note that for result (i) on mean adjacency matrix to hold, we only require the average maximum expected degree per layer $ \bar{\Delta} \gtrsim \log (2n/\epsilon)/M$. In comparison, a similar result for adjacency matrix of a single graph in \cite{chung11} (which (i) extends to multi-layer graph) requires the maximum expected degree $\Delta \gtrsim \log (2n/\epsilon)$. Hence the result holds for multi-layer graphs where the individual layers are sparser on average. At first glance the density condition on maximum expected degrees for result (ii) looks complicated. However, note that the condition is satisfied, for example, by a choice of $\Delta_{m} > \frac{4}{9} \log (2n/\epsilon)$ for each $m$ with $\epsilon=\frac{1}{Mn^2}$, which is the density condition for a similar result for the adjacency matrix of a single graph in \cite{chung11}. The condition as in (ii) relaxes the requirement that each layer of the multi-layer network be denser than a threshold, and hence one can have layers which are sparser as long as the layers together satisfy the density condition.

Over the next few sections we use the results of Theorem \ref{thm:lavrg} to prove consistency results for co-regularized spectral clustering, OLMF and spectral clustering in mean adjacency matrix. The common settings under which the results are proved are as follows. Let $\mathcal{G}$ be a multi-layer network with $M$ layers generated from the MLSBM with parameters $[Z,\mathbb{B}]$. Let $\mathbb{A}$ be its adjacency tensor. Let $\lambda^{(m)}$ denote the minimum in absolute value non-zero eigenvalue of the $m$th layer population adjacency matrix and $n_{\max}$ denote the number of nodes in the largest true community.

\subsection{Consistency result for co-regularized spectral clustering}
\begin{thm}
Let $[\hat{\mathbb{U}},\hat{U}^{*}]$ be the solution that maximizes the co-regularized spectral clustering objective function in (\ref{coreg}) applied to $\mathbb{A}$, and $r_{\text{coreg}}$ be the fraction of nodes misclusterd by a k-means procedure applied to $\hat{U}^{*}$. Assume $M\bar{\Delta} > \frac{4}{9} \log (4n/\epsilon)$ and all the $B^{(m)}$'s are of full rank. If we choose $\gamma_{m}$ large enough such that $\gamma_{m} >  \frac{\sqrt{M} \|A^{(m)}\|_2^2}{\sqrt{4 \bar{\Delta}\log (4n/\epsilon)}}$ for all $m$, then for any $\epsilon>0$, with probability at least $1-\epsilon$,
\[
r_{coreg}  \leq \frac{96n_{\max}k}{n\frac{1}{M}\sum_{m} \frac{(\lambda^{(m)})^2}{\Delta_m}} \sqrt{\frac{\bar{\Delta}\log (4n/\epsilon)}{M}}.
\]
\label{th:coreg}
\end{thm}

Several discussions on the results of Theorem \ref{th:coreg} are in order.
First, in the following lemma we replace the deterministic condition on $\gamma_{m}$ needed for consistency by a condition that holds only with high probability but involves quantities that depend purely on observed network statistics. Such a condition can then be easily verified in a given network.
\begin{lem}
Assume $M\bar{\Delta} > c \log (2n/\delta)$ where $c$ and $\delta$ are positive constants. For each $m$, if we choose each $\gamma_m > \frac{\sqrt{M} \|A^{(m)}\|_2^{2}}{\sqrt{\|\frac{2}{M}\sum_m A^{(m)}\|_2\log (4n/\epsilon)}}$, then for sufficiently large $c$, we have with probability at least $1-\delta$, $\gamma_{m} >  \frac{\sqrt{M} \|A^{(m)}\|_2^2}{\sqrt{4 \bar{\Delta}\log (4n/\epsilon)}}$ for all $m$.
\label{gammam}
\end{lem}

Although correct recovery under the noiseless case does not require any condition on $\gamma_{m}$'s, the consistency requires $\gamma_{m}$'s to be larger than a certain function of $\| A^{(m)}\|_2$ and $M$. In the typical case of sparse network layers that we will deal with, $\|A^{(m)}\|_2 \asymp \log n/M$, and then the condition in Lemma \ref{gammam} reduces to $\gamma_m > O( \|A^{(m)}\|_2)$. Based on this result, in our simulations in Section 5 we choose $\gamma^{(m)}$ to be a constant times $\max \|A^{(m)}\|_2$, identically in each layer.

Second, since it is not immediately clear when the above bound will imply consistent community detection, we make some further assumptions to simplify the bound. In particular we interpret the bound under a multi-layer extension of the four parameter stochastic blockmodel introduced in \citet{rcy11}.

\subsection*{Co-regularized spectral clustering under four parameter MLSBM}

We define a MLSBM of $M$ layers and $n$ nodes with four parameters $\mathbf{p}=\{p^{(1)},\ldots, p^{(M)}\},\mathbf{q}=\{q^{(1)},\ldots, q^{(M)}\},k,s$ as follows. In layer $m$, the connection probability within a community is $p^{(m)}$ and between communities is $q^{(m)}$. We assume $p^{(m)} \neq q^{(m)}$ but are of the same asymptotic order with respect to $n$, for all $m$. The number of communities is $k$ and all communities are of the same size $s=n/k$. Hence $n_{\max}=s=n/k$. We have the following lemma on the minimum eigenvalues of the population adjacency matrices $\lambda^{(m)}$'s.

\begin{lem}
For the four parameter MLSBM, $\lambda^{(m)}=s(p^{(m)}-q^{(m)})$, for all $m=1,\ldots,M $.
\label{eigenvalue}
\end{lem}

Let $a^{(m)} \frac{\Delta_{m}}{n}= p^{(m)}$ and $b^{(m)} \frac{\Delta_{m}}{n}= q^{(m)}$. Then $\lambda^{(m)}  = \frac{\Delta_{m}}{k} (a^{(m)}-b^{(m)})$. Consequently, the common asymptotic order of $p^{(m)}$ and $q^{(m)}$ is captured in the $\frac{\Delta_{m}}{n}$ term and $a^{(m)} \asymp b^{(m)} \asymp 1$. However, note that the difference $a^{(m)}-b^{(m)}$ could still be very small.  Define $f(\mathbf{a,b})=\frac{1}{M}\sum_{m} (a^{(m)}-b^{(m)})^2$. Then Theorem 2 implies
\begin{align*}
r_{coreg}  & \lesssim \frac{\frac{n}{k}k}{ n\frac{1}{M}\sum_{m}\frac{\Delta_m(a^{(m)}-b^{(m)})^2}{k^2}} \sqrt{\frac{\bar{\Delta}\log (4n/\epsilon)}{M}} \\
& \asymp \frac{k^2}{\frac{1}{M}\sum_{m}\Delta_m(a^{(m)}-b^{(m)})^2}\sqrt{\frac{\bar{\Delta}\log (4n/\epsilon)}{M}}.
\end{align*}
At this point we make a further assumption that $\Delta_m \asymp \bar{\Delta}$ for all $m$. Then we have
\[
r_{coreg} \lesssim \frac{k^2 \sqrt{\log (4n/\epsilon)}}{\sqrt{M\bar{\Delta}}f(\mathbf{a,b})},
\]
and community detection using this method is consistent as long as
$k=o((M\bar{\Delta}/
\log (4n/\epsilon))^{1/4}\sqrt{f(\mathbf{a,b})})$. We also note that the upper bound on misclustering rate becomes smaller as the number of layers $M$, the average density of the layers $\bar{\Delta}$ and a measure of community signal $f(\mathbf{a,b})$ increase.

We consider three growth regimes on the density of the component layers of the multi-layer graph. In the first regime we assume the dense graph setting where the vectors $\mathbf{p}$ and $\mathbf{q}$ do not change with $n$. This implies that $\bar{\Delta} \asymp n$ and consequently
\[
r_{coreg}  \lesssim  \frac{k^2}{\sqrt{nM/\log (4n/\epsilon)}f(\mathbf{a,b})}.
\]
Hence as long as $k=o((nM/\log (4n/\epsilon))^{1/4}\sqrt{f(\textbf{a,b})})$, $r_{coreg}\rightarrow 0$ with probability at least $1-\epsilon$, and we have consistent community detection.

In the second regime, we assume a semi-sparse setting where both $p^{(m)}$ and $q^{(m)}$ are of the order of $\log n/n$ for all $m$. Then $\bar{\Delta} \asymp \log n$ and  we have
\[
r_{coreg}  \lesssim \frac{k^2}{ f(\mathbf{a,b}) } \sqrt{\frac{\log (2n/\epsilon)}{M \log n}} \asymp \frac{k^2}{\sqrt{M} f(\mathbf{a,b})}.
\]
This implies that in this setting, as long as $k=o\left(M^{1/4} \sqrt{f(\mathbf{a,b})}\right)$, $r_{coreg}\rightarrow 0$, and we have consistent community detection.

Finally in the sparse ``constant degree" regime, where $p^{(m)}$ and $q^{(m)}$ are of the order of $1/n$ for all $m$, we have $\Delta_{m} \asymp 1$. Note that the density condition on the layers of the network for Theorem 2(i) to hold is $M \bar{\Delta} \gtrsim  \log n$, which can be satisfied even in the constant degree regime if $M \gtrsim \log n$. If this is satisfied, then we have from Theorem 2 that
\[
r_{coreg}  \lesssim \frac{k^2}{ f(\mathbf{a,b})} \sqrt{\frac{\log (2n/\epsilon)}{M}} \asymp \frac{k^2}{\sqrt{M/\log (2n/\epsilon)}f(\mathbf{a,b})}.
\]
Hence consistent community detection is possible as long as $k=o((M/\log (2n/\epsilon))^{1/4} \allowbreak \sqrt{f(\mathbf{a,b})})$. Consequently, a large number of very sparse graphs can also lead to consistent community detection, whereas in single layer networks consistent recovery is not possible in the constant degree regime. This is also true for spectral clustering in mean adjacency matrix as we will see in Theorem \ref{meangraph}, and is along the lines of the results obtained in \citet{pc15}.

The next section develops similar results for the OLMF method.

\subsection{Consistency result for orthogonal linked matrix factorization}

\begin{thm}
Let $[\hat{P},(\hat{\Lambda}^{(1)},\ldots, \hat{\Lambda}^{(M)})]$ be the solution that minimizes the OLMF objective function in (\ref{linked}) applied to $\mathbb{A}$, and $r_{LMF}$ be the fraction of nodes misclustered by a k-means procedure applied to $\hat{P}$. If the assumption in part (ii) of Theorem \ref{thm:lavrg} holds and at least one of the $B^{(m)}$'s is of full rank, then with probability at least $1-o(1)$,
\[
r_{OLMF} \leq \frac{48n_{\max}k\bar{\Delta}^{'1/2}(\bar{\Delta}^{1/2} + \bar{\Delta}'^{1/4}(\log 2M)^{1/2}(\log n)^{2+\epsilon}/M^{1/4} )}{\frac{1}{M}\sum_{m} (\lambda^{(m)})^2 n}.
\]
\label{th:lmf}
\end{thm}

This bound can also be simplified under the four parameter MLSBM defined earlier. Under the four parameter MLSBM with $\Delta_{m}$'s all being of the same order, we have $\Delta_m \asymp \bar{\Delta}$ and $\bar{\Delta}' \asymp \bar{\Delta}^2$. Then the bound in Theorem \ref{th:lmf}  simplifies to
 \begin{align*}
r_{LMF} & \lesssim \frac{\bar{\Delta}^{3/2} + \bar{\Delta}^{3/2}(\log 2M)^{1/2}(\log n)^{2+\epsilon}/M^{1/4} )}{ \frac{1}{M}\sum_{m}\frac{\Delta_m^2 (a^{(m)}-b^{(m)})^2}{k^2}} \\
& \asymp \frac{k^2 (1+(\log 2M)^{1/2}(\log n)^{2+\epsilon}/M^{1/4})}{\bar{\Delta}^{1/2}f(\mathbf{a,b})}\\
& \asymp \max \Bigg\{  \frac{k^2}{\bar{\Delta}^{1/2}f(\mathbf{a,b})},  \frac{k^2 ((\log 2M)^{1/2}(\log n)^{2+\epsilon})}{M^{1/4}\bar{\Delta}^{1/2}f(\mathbf{a,b})}\Bigg \}.
\end{align*}
In the dense case where $p^{(m)}$'s and $q^{(m)}$'s remain constant with increasing $n$, $\bar{\Delta} \asymp n$ and \[
r_{LMF} \lesssim \frac{k^2}{\min \{n^{1/2}f(\mathbf{a,b}), (M/(\log 2M)^2)^{1/4}(n/(\log n)^6)^{1/2}f(\mathbf{a,b})\}}.
\]
Hence consistent estimation is possible as long as $k$ grows slower than the square root of the term in the denominator.

In the sparser case of $\bar{\Delta} \asymp O( \log n)$ we similarly have
\[
r_{LMF} \lesssim  \max \Bigg\{  \frac{k^2}{(\log n)^{1/2}f(\mathbf{a,b})},  \frac{k^2 ((\log 2M)^{1/2}(\log n)^{3/2+\epsilon})}{M^{1/4}f(\mathbf{a,b})}\Bigg \},
\]
and consistency for the OLMF method provided
\[k=o\left (\min \Big\{(\log n)^{1/4},\frac{M^{1/8}}{(\log 2M)^{1/4}(\log n)^{3/4+\epsilon}}\Big\}\sqrt{f(\mathbf{a,b})}\right).
\]

\subsection{Consistency results for mean adjacency matrix}
The final result we prove provides an upper bound on the mis-clustering rate for consensus community detection using the usual single layer spectral clustering on the mean adjacency matrix.

\begin{thm}
 Define $\bar{A}=\frac{1}{M}\sum_{m=1}^{M}A^{(m)}$ and let $\lambda^{\bar{\mathcal{A}}}$ denote the minimum in absolute value non-zero eigenvalue of the mean population adjacency matrix $\bar{\mathcal{A}}=\frac{1}{M}\sum_{m=1}^{M}\mathcal{A}^{(m)}$. Let $r_{\text{av}}$ be the fraction of nodes misclustered by the spectral clustering algorithm applied to $\bar{A}$. If $M\bar{\Delta} >\frac{4}{9}\log (2n/\epsilon)$, and $\bar{B}=\frac{1}{M}\sum_{m=1}^{M}B^{(m)}$ is of full rank (i.e., rank $k$), then with probability at least $1-\epsilon$,
\[
r_{av} \leq \frac{256 n_{\max}k \bar{\Delta}\log (2n/ \epsilon) }{(\lambda^{\bar{\mathcal{A}}})^2nM}.
\]
\label{meangraph}
\end{thm}

To prove this result, we employ a proof technique using Theorem \ref{thm:lavrg}, which is different from \citet{hxa14} and allows us to characterize the dependence of the misclustering rate on the growth rates of various MLSBM parameters. While the concentration result in Frobenius norm of \citet{hxa14} would imply consistent community detection through spectral clustering in mean adjacency matrix for fixed $k$ as long as $n=o(M^{1/2})$, our technique yields a non-asymptotic bound on the mis-clustering rate with direct dependence on the number of communities, sparsity, signal to noise ratio along with $n$ and $M$. We will next analyze the non-asymptotic bound under a simplified model and different asymptotic growth criteria on the above quantities.

Note the presence of $\lambda^{\bar{\mathcal{A}}}$ in the denominator of the bound implies that the bound depends on the eigen-gap of the mean adjacency matrix. To interpret the bound under the four parameter MLSBM, we first prove the following lemma on the eigen-gap $\lambda^{\bar{\mathcal{A}}}$.

\begin{lem}
For the four parameter MLSBM,
$\lambda^{\bar{\mathcal{A}}}=s\frac{1}{M}\sum_{m} (p^{(m)}-q^{(m)})$.
\label{meaneigenvalue}
\end{lem}
Similar to previous cases, writing the result in terms of $\bar{\Delta}, a^{(m)}$, and $b^{(m)}$ we have $\lambda^{\bar{\mathcal{A}}}=\frac{1}{M}\sum_{m} \frac{\Delta_m}{k} (a^{(m)}-b^{(m)}) \asymp \frac{\bar{\Delta}}{k}\frac{1}{M}\sum_{m}  (a^{(m)}-b^{(m)})$. Define $g(\mathbf{a,b})=(\frac{1}{M}\sum_{m} (a^{(m)}-b^{(m)}))^2$. Then from Theorem \ref{meangraph} we have with probability at least $1-\epsilon$,
 \[
 r_{av} \lesssim \frac{\frac{n}{k}k \bar{\Delta}\log (2n/ \epsilon) }{(\frac{\bar{\Delta}}{k})^2 g(\mathbf{a,b})nM } \asymp \frac{k^2}{M\bar{\Delta}g(\mathbf{a,b})/\log (2n/ \epsilon) }.
 \]

This implies that $ r_{av} \rightarrow 0$ as long as $k=o(\sqrt{M\bar{\Delta}g(\mathbf{a,b})/\log (2n/ \epsilon)})$, and we have consistent community detection. We note that $g(\mathbf{a,b})$ and $\bar{\Delta}$ are averages over the layers of the corresponding quantities for single layer case. The above result then implies that with increasing number of layers $M$, the upper bound on the misclustering rate gets smaller by a factor of $\sqrt{M}$ as compared to applying spectral clustering on any one of the layers separately as shown in \citet{qr13} and \citet{lei14} (The $\log n$ term does not appear in \citet{lei14} due to tighter bound on $\|A-\mathcal{A}\|_2$). We also note that the denominator in the rate for $r_{av}$ contains the term $g(\mathbf{a,b})=(\frac{1}{M}\sum_{m} (a^{(m)}-b^{(m)}))^2$ instead of $f(\mathbf{a,b})=\frac{1}{M}\sum_{m} (a^{(m)}-b^{(m)})^2$, which appeared earlier in the rates of OLMF and co-regularized spectral clustering. From Jensen's inequality,
\[
g(\mathbf{a,b})=(\frac{1}{M}\sum_{m} (a^{(m)}-b^{(m)}))^2 \leq \frac{1}{M}\sum_{m} (a^{(m)}-b^{(m)})^2=f(\mathbf{a,b}),
\]
with equality holding if and only if all the $(a^{(m)}-b^{(m)})$'s are equal. Hence equality holds if the layers are of similar signal quality, and otherwise $f(\mathbf{a,b})$ is larger than $g(\mathbf{a,b})$. Hence the goodness of the rate for spectral clustering in mean adjacency matrix depends on if the aggregate of the layers has good signal quality or not. In the situation where some of the layers in the multi-layer network contain heterophilic communities while the others contain homophilic communities, then $a^{(m)}-b^{(m)}$ is negative in some layers and positive in other layers. In that case $\lambda^{\bar{\mathcal{A}}}$ could be very small and performance guarantee on spectral clustering of mean adjacency matrix become poor. These conclusions are in line with previous conclusions from minimax rates and phase transitions of consistency thresholds in \citep{pc15}.

In the dense regime where the vectors $\mathbf{p}$ and $\mathbf{q}$ do not change with $n$, we have the mis-clustering rate in spectral clustering in mean adjacency matrix is bounded by $
 r_{av} \lesssim \frac{k^2}{nM g(\mathbf{a,b})/\log (2n/\epsilon)}.$
In the semi-sparse regime where both $p^{(m)}$ and $q^{(m)}$ are of the order of $\frac{\log n}{n}$ for all $m$, we have $\bar{\Delta} \asymp \log n$ and $r_{av} \lesssim \frac{k^2}{M g(\mathbf{a,b})}$.
Finally, in the sparse constant degree regime where both $p^{(m)}$ and $q^{(m)}$ are of the order of $1/n$ for all $m$, we have $
 r_{av} \lesssim \frac{k^2}{M g(\mathbf{a,b})/\log (2n/ \epsilon) }.$

\section{Simulation studies}
In this section, we numerically compare the performance of the following methods through a principled simulation study: spectral clustering on mean adjacency matrix (Mean adj.), OLMF, co-regularized spectral clustering (Coreg spec), spectral clustering on aggregate spectral kernel (SpecK) and the module allegiance matrix (Module alleg.). Since the computational algorithms for both OLMF and Coreg Spec are only expected to reach a local optimum, it is important to supply good initial conditions to them and also take the best solution based on multiple initial conditions. We initialize the OLMF algorithm with $P$ being the community assignment matrix from a randomly chosen layer and $\Lambda^{(m)}$ being the matrix containing the top $k$ eigenvalues of $A^{(m)}$ in the diagonal. For the co-regularized spectral clustering algorithm we choose $\gamma^{(m)}$ as $4 \max \|A^{(m)}\|_2$ for all $m$, since the theoretical results have indicated that $\gamma^{(m)}$ should be larger than $\|A^{(m)}\|_2$ for each $m$.

For the first three simulations, we simulate networks from the MLSBM with the number of nodes $n=600$ and the number of layers $M=5$, under three different scenarios on the connection probability matrices of different layers. The performances of the methods are evaluated with increasing average degree of the multi-layer network since we would expect any reasonable method to perform better as the network gets denser. The number of communities is fixed at $3$ and we assume it to be known in advance. The fourth simulation involves generating networks from MLSBM with varying number of layers and testing the performance of the methods with increasing number of layers. The fifth and final simulation considers the scenario where the multi-layer network contains layers with both heterophilic and homophilic communities.

The evaluation criterion is the normalized mutual information (NMI) with the ground truth community assignments which generate the network. The NMI is an information theoretic measure of similarity between two vectors of community assignments, with $1$ indicating a perfect match and $0$ indicating the vectors are random with respect to each other. The first three experiments are replicated 40 times while the last two experiments are repeated 100 times, and the average performance across the repetitions is reported.

The data are generated according to MLSBM as defined in (\ref{SBM}) in the following fashion. The community vector $Z_{i}$ for each node $i$ is generated according to a multinomial distribution with equal probability of being in any of the 3 communities. The block model matrices $B^{(m)}$'s in different layers are generated by the following scheme. Let $\delta$ be the vector of $k$ diagonal elements and $\epsilon$ be the vector of $k^2-k$ off-diagonal elements. We generate half of the elements of the $\epsilon$ vector from a uniform distribution $U(a,b)$ within a short range $[a,b]$ and the other half is a replication of the first half such that the matrix is symmetric. The elements of $\delta$ are generated from $U(\rho a,\rho b)$, where $\rho$ is the parameter that controls the signal to noise ratio (SNR). We call an SNR of 2-3 as ``strong" signal and an SNR which is only slightly greater than $1$ as ``weak" signal.

\begin{figure}[h]
\centering{}
\begin{subfigure}{0.45\textwidth}
\centering{}
\includegraphics[width=.95\linewidth]{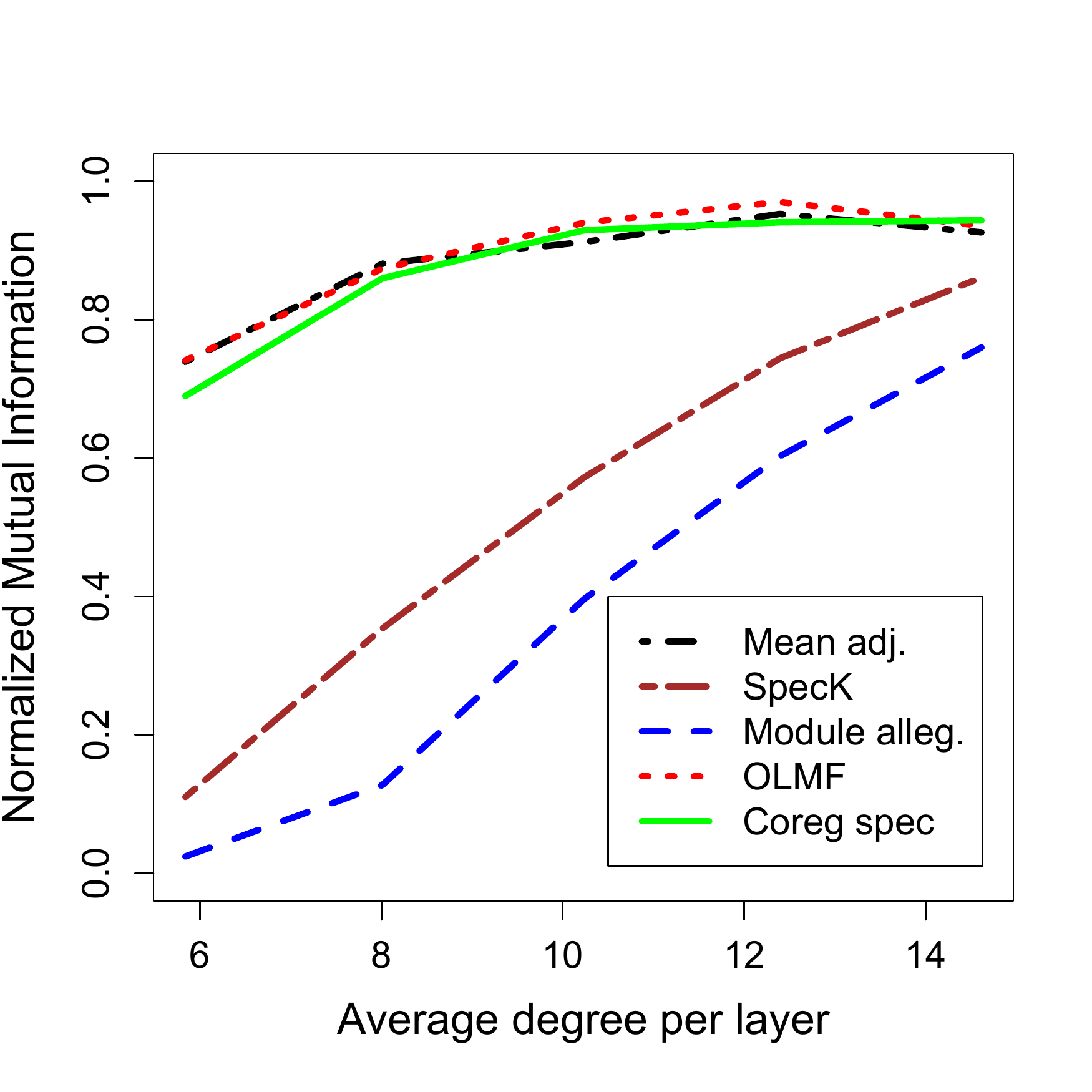}
\end{subfigure}%
\begin{subfigure}{0.45\textwidth}
\centering{}
\includegraphics[width=.95\linewidth]{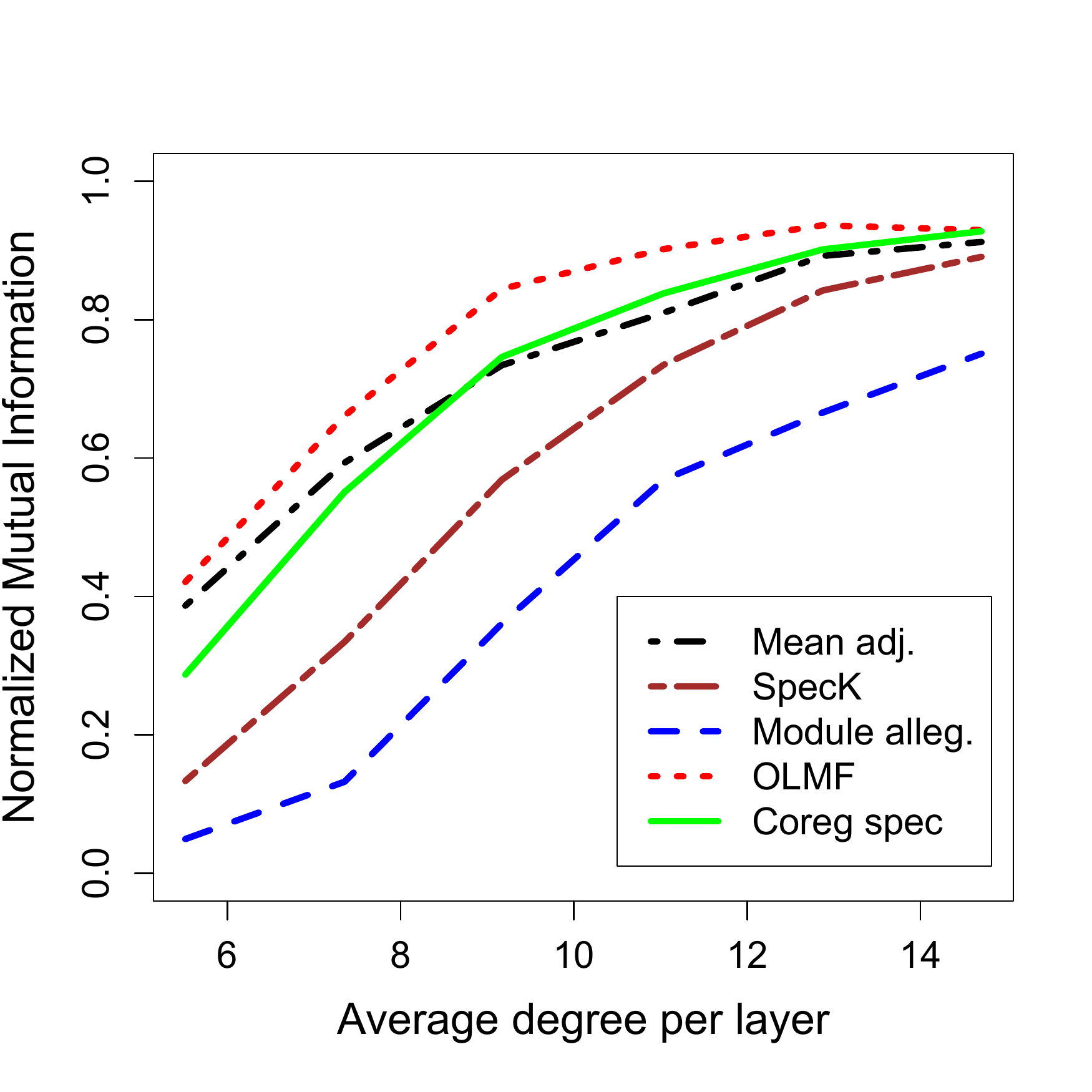}
\end{subfigure}
\begin{center} (a) \hspace{160pt} (b) \end{center}
\caption{Performance of various methods with increasing average degree of nodes for data generated from MLSBM with $600$ nodes, $5$ layers and $3$ communities. (a) All layers have strong signals with some variations; (b) the layers are mixed in terms of signal quality.}
\label{SM}
\end{figure}

\subsection*{Strong community signals}
In the first simulation from MLSBM, we make all the layers contain generally strong signals, but the exact SNR is randomly varied slightly so as to have some variations in signal quality across the layers. The performance of various methods under consideration is presented in Figure \ref{SM}(a). Note that the layers are sparse at an average initially which is evident from the low average degree per layer: an average degree of 6 in a layer of 600 nodes, which is about 1\% degree density. The layers then become denser gradually and reach about 2.5\% degree density per layer. The performance of all the methods generally increases with increasing average degree. We note that spectral clustering on mean adjacency matrix, OLMF and co-regularized spectral clustering perform similarly throughout the range of the simulation. The aggregate of spectral kernel and module allegiance matrix method substantially underperform, especially in sparse multi-layer networks.

\subsection*{Mixed and ambiguous community signals}
In this simulation, the component layers are mixed in community signal quality in the following manner. We have three layers with strong community signals and two layers where the community structure is ambiguous or almost non-existent due to weak signal to noise ratio. This scenario is very useful to test the robustness of methods against possible variation or absence of community patterns in some of the layers. The results are presented in Figure \ref{SM}(b). The OLMF method performs the best over the entire range of values of average degree, followed by co-regularized spectral clustering and spectral clustering of mean adjacency matrix. The aggregate spectral kernel and module allegiance matrix methods once again perform poorly when the average density in the layers is low, but recover subsequently as the layers become denser. The spectral kernel method performs better than the module allegiance matrix method in both the strong signals and mixed signals scenarios.

\begin{figure}[h]
\centering{}
\begin{subfigure}{0.33\textwidth}
\centering{}
\includegraphics[width=.95\linewidth]{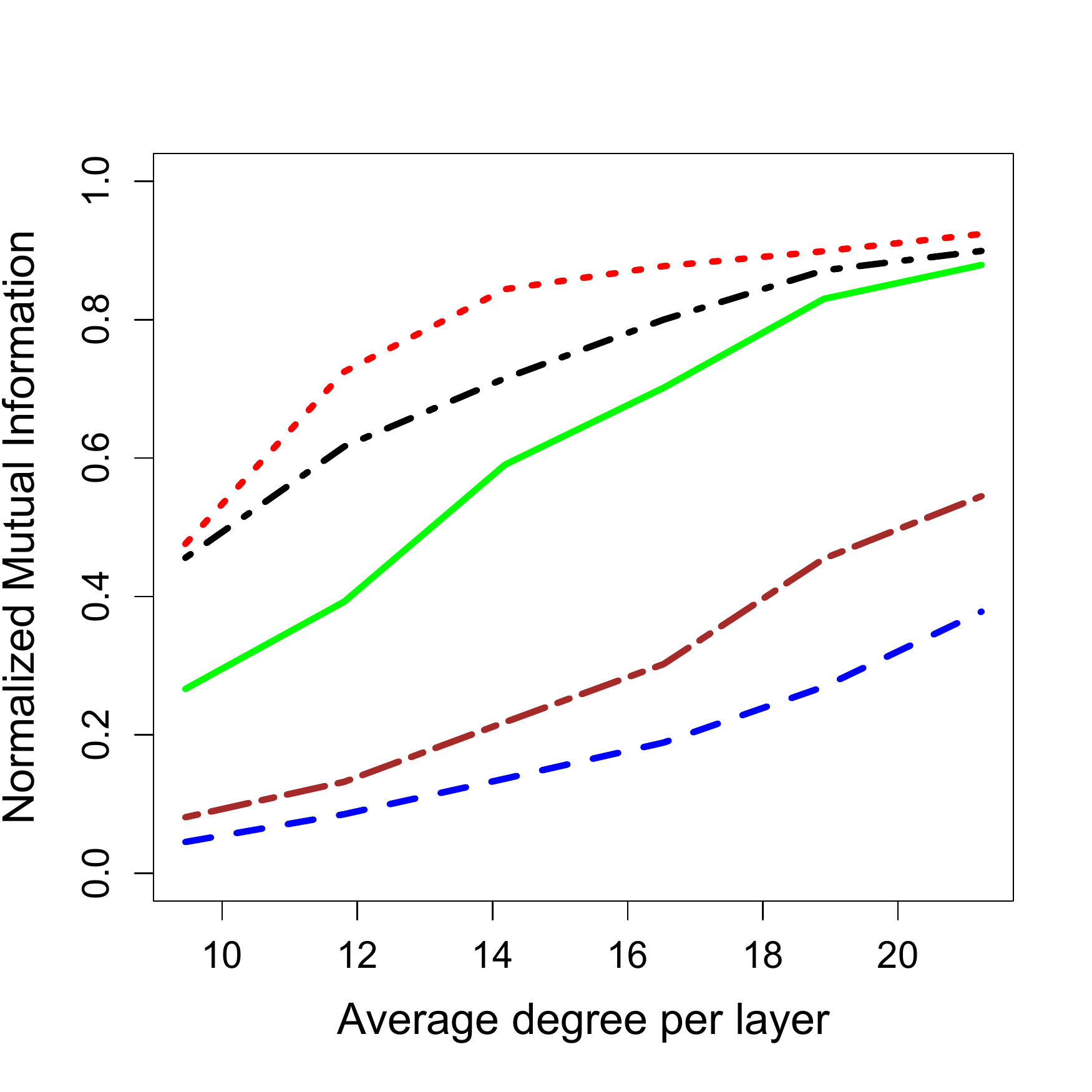}
\end{subfigure}%
\begin{subfigure}{0.33\textwidth}
\centering{}
\includegraphics[width=.95\linewidth]{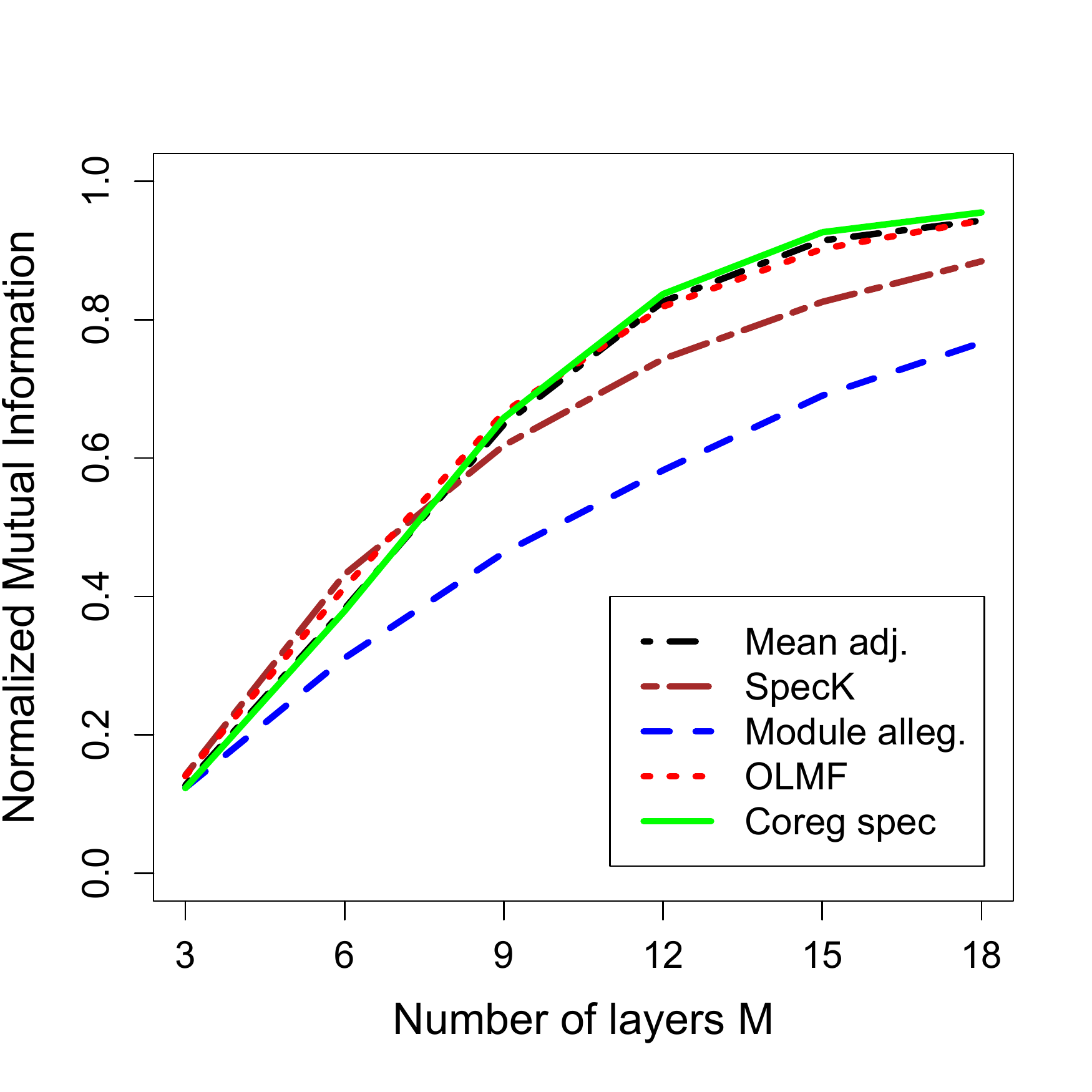}
\end{subfigure}%
\begin{subfigure}{0.33\textwidth}
\centering{}
\includegraphics[width=.95\linewidth]{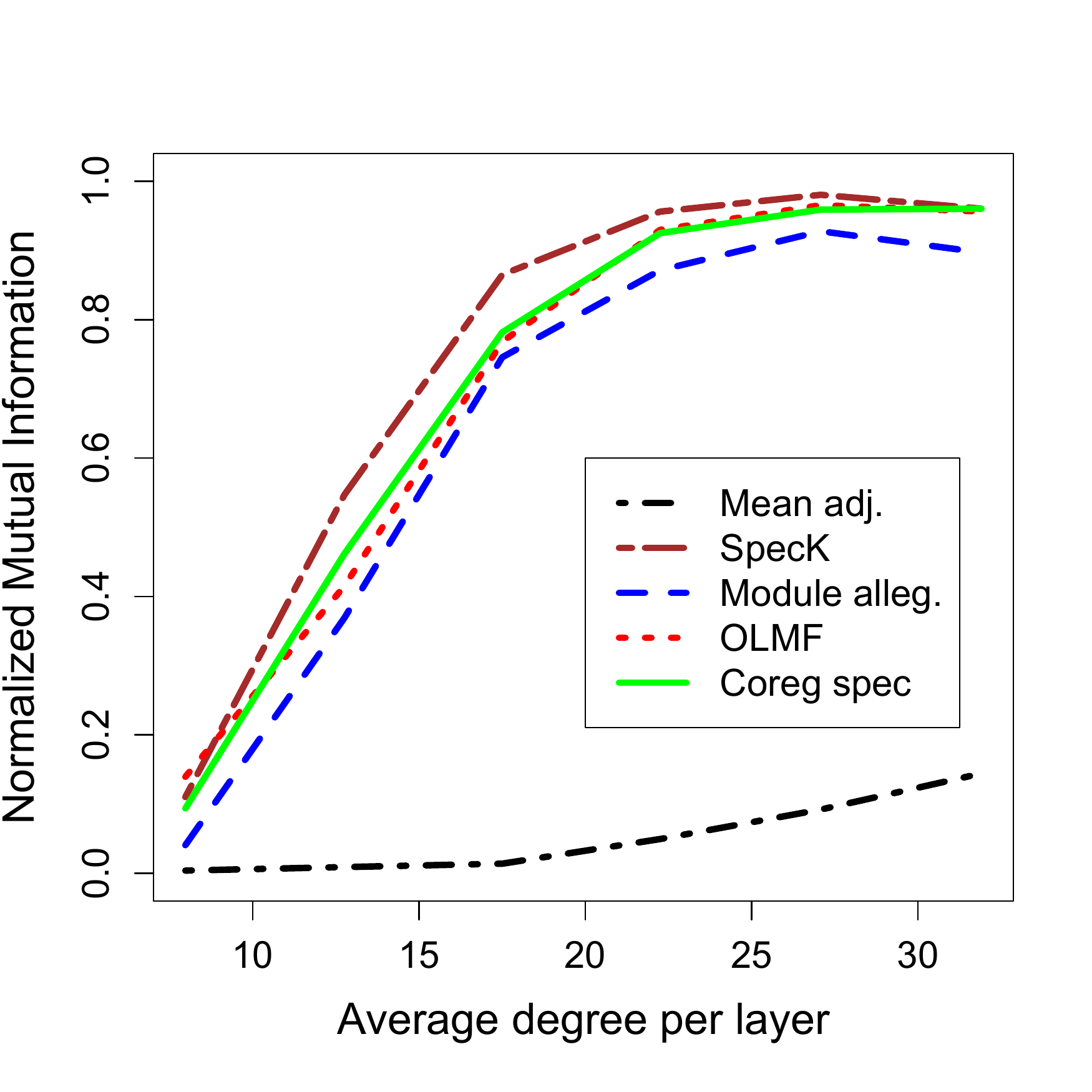}
\end{subfigure}
\begin{center} (a) \hspace{100pt} (b) \hspace{100pt} (c)\end{center}
\caption{Performance of various methods with (a) increasing average degree of nodes for data generated from MLSBM with $600$ nodes, $5$ layers and $3$ communities, (b) increasing number of layers with $300$ nodes and $6$ communities, (c) increasing average degree of nodes with $600$ nodes, $5$ layers and $3$ communities, where 3 layers contain homophilic communities and the other 2 contain heterophilic communities. The labels in (b) and (c) are shared for all figures.}
\label{CI}
\end{figure}

\subsection*{Complementary information}
The third scenario considers the so-called ``complementary" principle of multiple views in multi-view learning \citep{liu13}. In our case, this is equivalent to the following: none of the layers alone is sufficient to describe the community structure properly, but the layers can complement each other and together describe the community patterns. For our simulation, we generate data from MLSBM with 600 nodes, 5 layers and 3 communities with the following setting. In each of the first 3 layers, two of the communities are difficult to distinguish from noise while the third community has a SNR of 3. The fourth layer has two of the communities with high SNR and the fifth layer has the same two communities with low SNR. The performance of the competing methods are presented in Figure \ref{CI}(a). We observe that both aggregate of spectral kernel and module allegiance matrix method perform poorly in this scenario as compared to the intermediate fusion methods as well as spectral clustering of mean adjacency matrix. This is expected since none of the layers alone contain complete information about the community structure and hence the eigenspaces computed separately are not very informative of the community structure. Consequently sharing information while computing individual eigenspaces as well as a consensus eigenspace is beneficial as opposed to a late fusion of individual eigenspaces. In addition, the OLMF method appears to have a clear advantage in this scenario over both co-regularized spectral clustering and spectral clustering of the mean adjacency matrix.

\subsection*{Increasing number of layers}
This simulation setup tests the abilities of the methods to recover the community structure with a small fixed number of nodes, but increasing number of layers (and consequently more data). However, as is the case with many real world multi-layer networks, not all of the layers are strongly informative of the community structure. We fix $n$ at 300, $k$ at 6 and increase $M$ from 3 to 18 in steps of 3. At every step, we add 3 layers to the multi-layer network, two of which have weak signal quality, while the third one has a strong signal. The performance of the competing methods in this simulation with 100 repetitions is depicted in Figure \ref{CI}(b). We observe that the accuracy of consensus community detection in all the methods generally increases with increasing number of layers. As with the previous scenarios, we observe that OLMF, co-regularized spectral clustering, and spectral clustering of the mean adjacency matrix have more improvement in performance as compared to aggregate of spectral kernel and module allegiance matrix methods.

\subsection*{Layers with heterophilic communities}

Finally, we consider the scenario where some layers contain homophilic (assortative) communities while others contain heterophilic (disassortative) communities. The layers with heterophilic communities have less density within the blocks as opposed to inter-block densities. Such interactions with disassortative communities are commonly encountered in food webs and social networks.
From our theoretical analysis we expect the spectral clustering of mean adjacency matrix to perform poorly in this setting. Intuitively, the mean adjacency matrix has strong inter-community edge density (in addition to strong intra-community edge density) due to the layers with heterophilic communities and consequently, the community structure  is ambiguous and difficult to detect. However, the community information is separately available in all the layers irrespective of whether the communities in that layer are homophilic or heterophilic. Then one would hope, perhaps, a different way of combining information from layers will yield the community structure correctly.

Since the inter-block connection probabilities are higher than intra-block connection probabilities in the layers with heterophilic communities, the eigenvalues corresponding to the eigenvectors that contain the clustering information are all negative. Hence we need to modify some of the methods slightly for this scenario. For aggregate spectral kernel and module allegiance methods, we choose the eigenvectors corresponding to the top $k$ eigenvalues in \textit{absolute value} to form the $\hat{U}^{(m)}$ matrix in each layer. For co-regularized spectral clustering, we update the $\hat{U}^{(m)}$ matrix during the alternating eigen-decomposition by selecting the vectors corresponding to the top $k$ eigenvalues in \textit{absolute value}. The mean adjacency matrix and OLMF methods do not require any change to be made, however, we make the following optional modifications. For mean adjacency matrix during the eigen-decomposition, we choose eigenvectors corresponding to the top $k$ eigenvalues in \textit{absolute value} of the mean adjacency matrix, while for OLMF we only change the initialization of $\Lambda^{(m)}$ matrices to include the $k$ largest eigenvalues in \textit{absolute value} as its diagonal.

We fix $n$ at 600, $k$ at 3, $M$ at 5, and increase the average degree per layer from 8 to 32 (from about 1\% to 4.5\% in degree density). We make 3 of the 5 layers contain homophilic communities by setting the $\rho$ parameter (SNR) at 3, while we make the other two layers contain heterophilic communities by setting $\rho=1/3$ so that the elements of $\delta$ are smaller than that of $\epsilon$. The results are presented in Figure \ref{CI}(c).

As expected from our theoretical results, we observe that the performance of spectral clustering in mean adjacency matrix completely breaks down and is substantially worse than the competing methods in this scenario.
The other four methods behave similarly and the accuracy of community detection steadily increases with increasing degree density. This indicates that all of those four methods are capable of extracting information relevant to the community structure from layers with both homophilic and heterophilic communities and combine them without nullifying the information.

We also note that the aggregate spectral kernel method performs slightly better compared to the two intermediate fusion methods throughout the range of the simulation. We think this is because of the following reason.
The relatively higher average degree per layer in the simulated networks compared to, e.g., that in Figure \ref{SM}(a), means the recovery of the true eigen-spaces (which contain the information on community assignments) by spectral methods in each of the layers becomes increasingly accurate irrespective of whether the communities are homophilic or heterophilic \cite{rcy11,lei14,qr13}. This leads to better performance of the methods that purely rely on combinations of those independently obtained eigen-spaces. Hence the aggregate spectral kernel itself becomes more effective than the intermediate fusion methods. The intermediate fusion methods on the other hand, shares information while computing the eigen-spaces and the consensus eigen-space appears to underperform in the presence of layers with both homophilic and heterophilic communities.

\subsection*{Discussion on the simulation results}
Our simulations clearly show that in sparse networks the intermediate fusion of information based methods, OLMF and co-regularized spectral clustering, perform better than late fusion methods, aggregate of spectral kernel and module allegiance matrix method. We think sharing information across layers while computing individual layer wise  spectral embeddings increases the accuracy in each of them, and hence the centroid is a more effective combination than aggregate spectral kernel or module allegiance matrix type of combination. The spectral clustering of mean adjacency matrix performs well in our simulations except the last scenario where the multi-layer network contains layers with both homophilic and heterophilic communities, in which case its performance is extremely poor. We also observe in our simulations that aggregate spectral kernel performs better than module allegiance matrix. We think the performance in module allegiance suffers because of additional noise introduced in discrete community assignments. Overall, we think the intermediate fusion methods outperform or remain competitive to the baseline methods of aggregate spectral kernel and spectral clustering in mean adjacency matrix in a wide variety of scenarios.

\section{Conclusions and discussions}
In this paper we have analyzed a number of spectral and matrix factorization based techniques for multi-view clustering in terms of their asymptotic consistency properties for community detection in multi-layer networks generated from the MLSBM. We have considered a high dimensional asymptotic framework where both the number of layers ($M$) and the number of nodes ($n$) of the multi-layer graph grow.  The main technical contribution of the article is to prove non-asymptotic error bounds for community detection using the global optimal solutions of both co-regularized spectral clustering and OLMF, and spectral clustering of the mean adjacency matrix in terms of model parameters of the MLSBM.  As an intermediate step we have proved two concentration inequalities on two functions of adjacency matrices of a multi-layer network. We have further shown that the above-mentioned methods enjoy consistency guarantees under some conditions on the number of communities $k$, the maximum expected degrees of the layers $\Delta_m$'s and the signal to noise ratios of the layers.

We have also compared five methods in terms of finite sample performance under data generated from the MLSBM through a simulation study. We found both the co-regularized spectral clustering and OLMF to be robust under varied scenarios. We also note from the simulations that widely popular methods where each layer is dealt separately and the results are fused at a later state, such as aggregating spectral kernels or module allegiance matrix, do not perform well in sparse networks when the individual layers do not contain sufficient information to recover the community structure efficiently. However, the OLMF and co-regularized spectral clustering perform well in those scenarios. We hypothesize that this is due to sharing information across layers while computing the community structure solution at each layer.

\subsection*{Global optimizers}
Throughout the paper we have studied the properties of the global optimal solutions of the optimization problems under consideration. However, in the absence of computational methods guaranteed to achieve the global optimal solutions, it is not known whether such global optimum can ever be achieved under any circumstances. Indeed, the algorithms we have used to compute the solutions in our simulation studies are approximate algorithms that can at best reach a local optimum. To the best of our knowledge, no computationally feasible algorithm exist that can compute the global optima of the intermediate fusion objective functions with guarantees. We view the results obtained in this article as only a first step in the direction of understanding the behavior of multi-view learning methods in the context of community detection in multi-layer networks. In the future, we hope to investigate possibilities of obtaining algorithms with global optimum guarantees and extend the results obtained here to such cases.

To assure ourselves that the solutions computed by the algorithms used here are not completely away from the global solutions, we conducted a simulation study. Although we do not know the true global optimum of the two optimization problems under study in real data situations, or even in simulated sample networks, we know the solutions for them theoretically when they are applied to the population adjacency tensor. Hence we study the objective function values at convergence for the two methods applied to population adjacency tensors with increasing degree density. Our first simulation verified that using the spectral clustering of mean adjacency matrix as initial solution, the OLMF objective function is within $10^{-18}$ of 0 and the co-regularized spectral clustering objective function is within $10^{-12}$ of the true maximum across the range of the simulation. Our second simulation whose result is presented in Figure \ref{robust}(a), verified that starting from a random initial solution the objective function value for co-regularized spectral clustering goes close to the true maximum with increasing number of iterations eventually being equal to the true maximum.

\subsection*{Consensus community detection in the presence of noise}
In this paper we have assumed the presence of an underlying consensus community assignment for the multi-layer network and focused on the problem of detecting such a structure. Indeed, Lemma \ref{recoverySBM}, which shows that the methods under study can correctly recover the community structure from the population adjacency tensor (i.e., without sampling noise), is the crucial backbone of the paper on which all results are based on. Here we analyze a scenario where the community structure is truly present in some layers (perhaps a majority), while it is either absent or is different in the other layers, and the task is to detect the community structure present in the majority of the layers. Such scenarios have been previously considered in \cite{stanley15,wilson2016community}. Since this scenario is not the focus of the paper, we will primarily analyze whether the methods are capable of recovering the community structure from the population adjacency tensor of such a multi-layer SBM. Let $M_1$ layers contain the community structure of interest $Z_1$ and $M_2$ layers contain a different community structure $Z_2$ with $M_1>M_2$. The community structure $Z_2$ could put all vertices in the same community (i.e., simply an Erdos-Renyi graph) or could be a community structure that is different from $Z_1$. We concentrate on the former case, where $Z_2$ does not define any community structure. We will study under what conditions the methods analyzed in this paper will be able to detect the main community structure of interest $Z_1$. Even for the $M_2$ layers with no community structure, we can write the population adjacency matrices as still being created by $Z_1BZ_{1}^T$ but with $B$ having identical values in each entry and consequently of rank $1$.

The mean adjacency matrix can then be written as:
\[
\bar{\mathcal{A}}=\frac{1}{M}\sum_{m=1}^{M}\mathcal{A}^{(m)} = Z_1 \left(\frac{1}{M}(\sum_{m=1}^{M_1}B^{(m)}+M_2B')\right)Z_1^{T}.
\]
Nevertheless, we would require a similar condition as before, namely $\bar{B}=\frac{1}{M}(\sum_{m=1}^{M_1}B^{(m)}+M_2B')$ will be full rank (i.e., rank $k$). Under this condition the spectral clustering procedure in mean adjacency matrix can extract the true community structure from the population adjacency matrix.

An (simplified) extension of the four parameter MLSBM can be defined for this case as follows: let $a$ and $b$ be diagonal and off-diagonal elements of $B^{(m)}$'s in the first $M_1$ layers with $a>b$ and $c$'s are the elements of $B'$. From Lemma \ref{meaneigenvalue}, the smallest non-zero eigenvalue of $\bar{B}  = \frac{1}{M}(M_1(a-b)I_k + (M_1b+M_2c)1_k1_k^T) $ is $\frac{M_1}{M}(a-b)$. Since $a>b$, the $k \times k$ matrix $\bar{B}$ has $k$ non-zero eigenvalues and consequently is of full rank. Then a spectral clustering algorithm can recover the true community structure from the mean population adjacency matrix.

However, the noise plays a big role when we look at the sample adjacency matrices. Using the simplified result of Theorem \ref{meangraph} under the four parameter MLSBM, we have consistency as long as $M\bar{\Delta}g(a,b)/\log n = \omega (1)$ for a constant number of communities $k$. Now if we assume all layers are of similar density, then $\bar{\Delta}$ does not change by adding Erdos-Renyi graphs. However, $g(a,b)=\frac{M_1^2(a-b)^2}{(M_1+M_2)^2}$ decreases as we increase $M_2$. In the case of $\bar{\Delta} \asymp \log n$, we have $M_1(a-b)^2 =\omega(M)$ as a sufficient condition for consistency. This is in contrast to the usual requirement of $(a-b)^2 =\omega(1)$. On the other hand if we assume $M_1$ and $M_2$ are fixed, but the density of the Erdos-Renyi layers gradually increases, then with the addition of such dense but uninformative layers, $\bar{\Delta}$ increases, while $g(a,b)$ remains the same. This increases the upper bound and the method does not lead to consistent community detection anymore.

\begin{figure}[h]
\centering{}
\begin{subfigure}{0.38\textwidth}
\centering{}
\includegraphics[width=\linewidth]{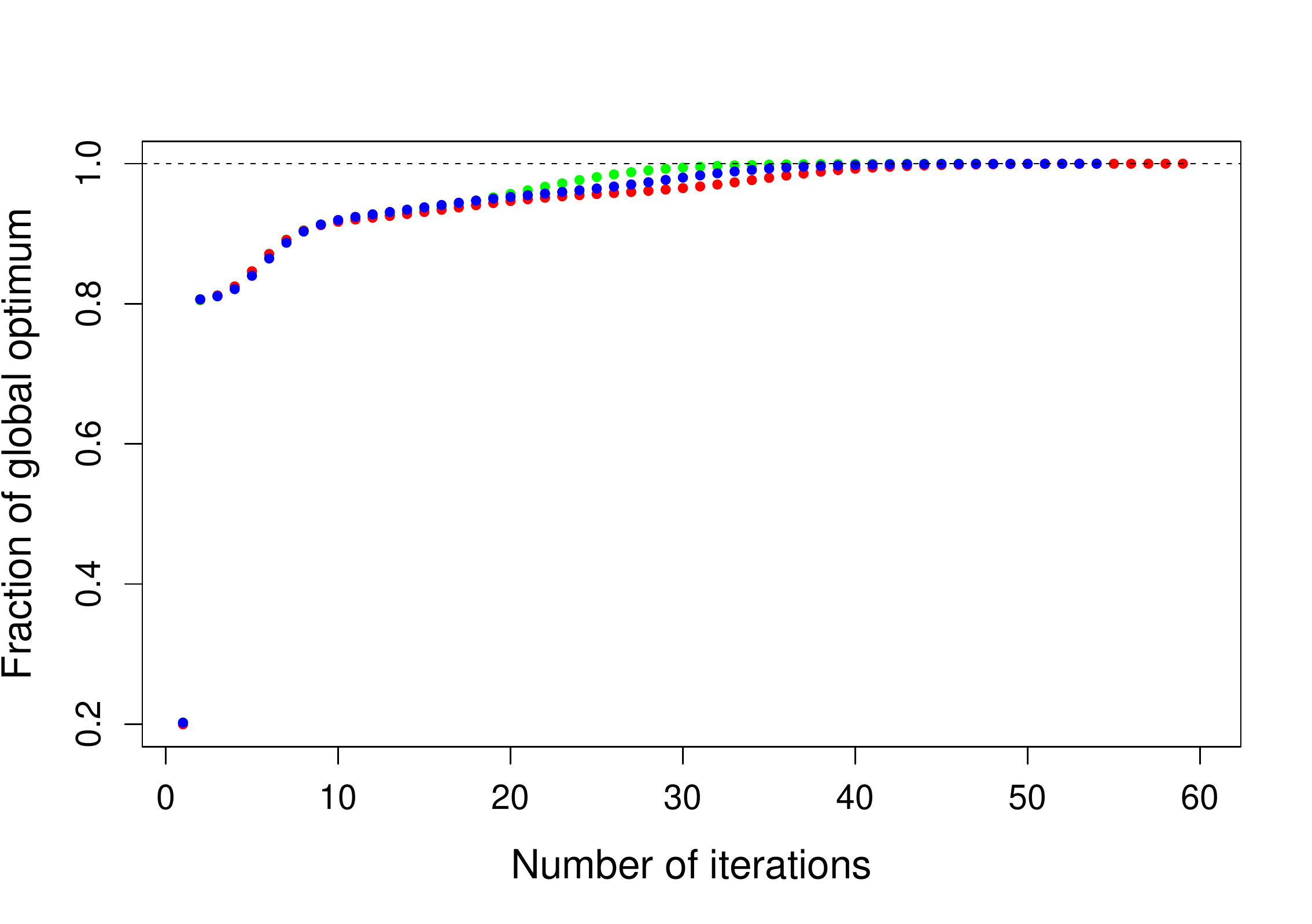}
\end{subfigure}%
\begin{subfigure}{0.32\textwidth}
\centering{}
\includegraphics[width=\linewidth]{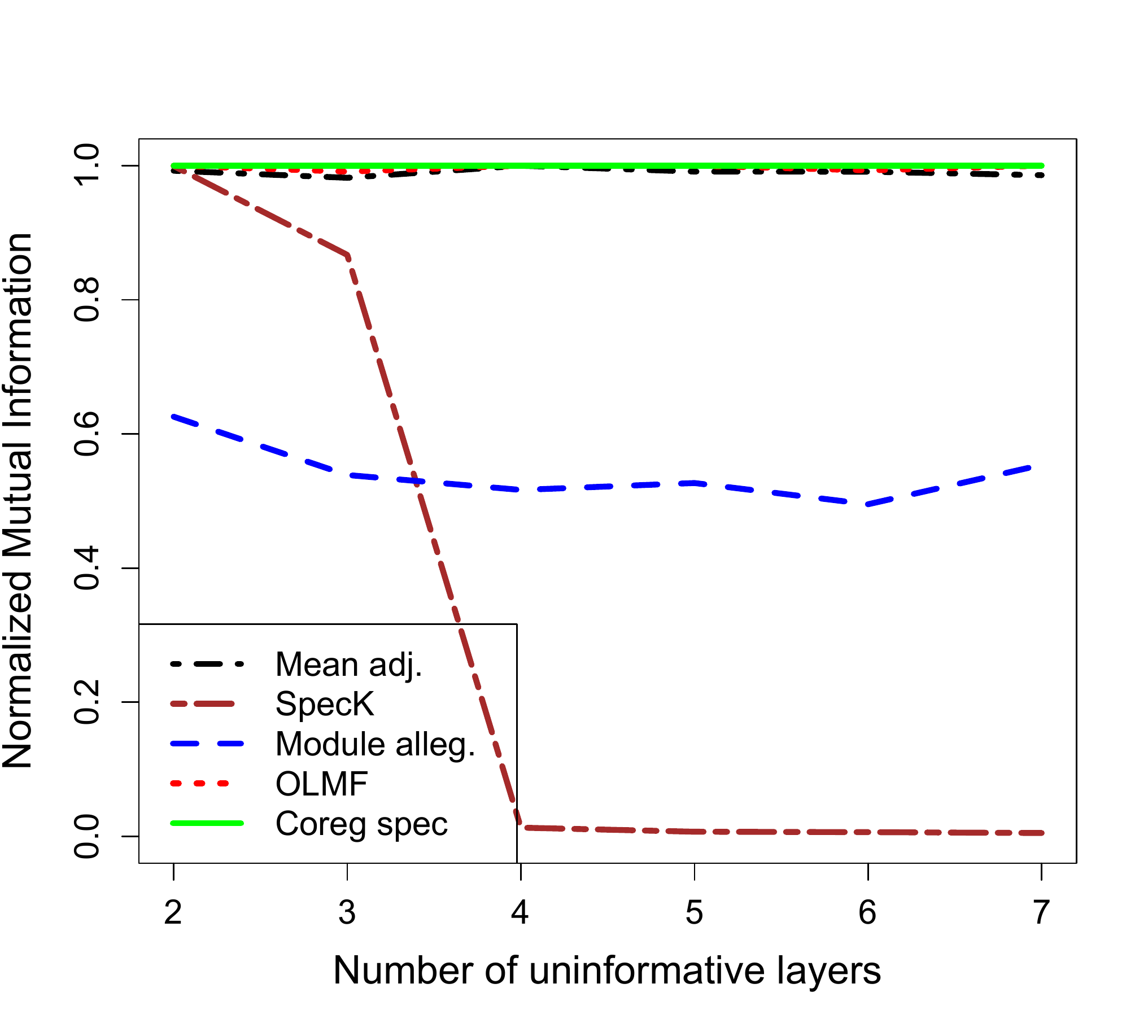}
\end{subfigure}%
\begin{subfigure}{0.30\textwidth}
\centering{}
\includegraphics[width=\linewidth]{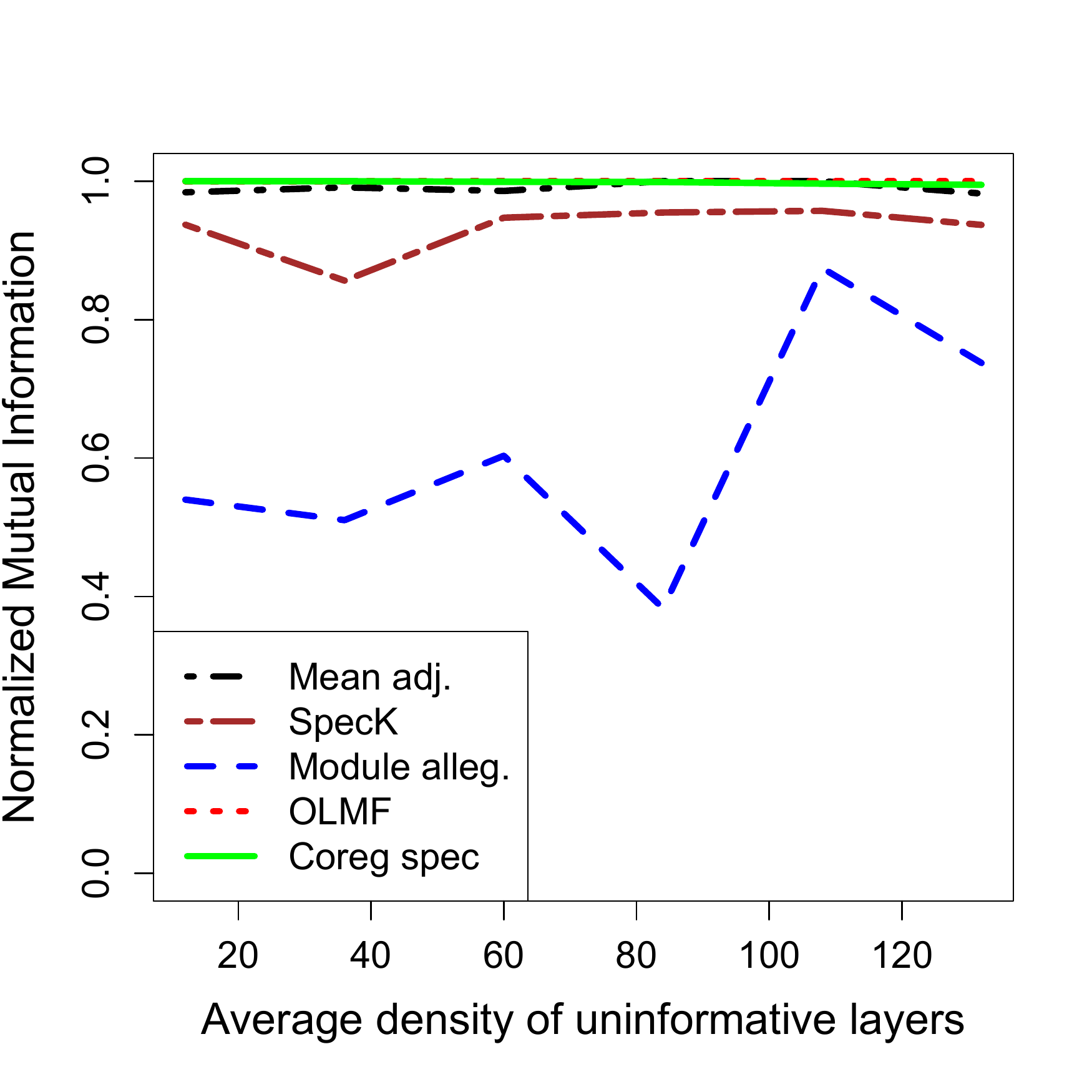}
\end{subfigure}
\begin{center} (a) \hspace{100pt} (b) \hspace{100pt} (c)\end{center}
\caption{(a) Objective function value as the fraction of the global optimum with increasing number of iterations for co-regularized spectral clustering. (b)-(c) Recovery of the true community structure from population adjacency tensor in the presence of uninformative layers: (b) The number of uninformative layers is increased keeping their densities fixed; (c) the density of the $3$ uninformative layers are gradually increased. In both (b) and (c) there are $3$ layers informative of the community structure.}
\label{robust}
\end{figure}

Turning our attention to spectral kernel method, we will have $Z_1^T(Z_1^TZ_1)^{-1}Z_1^T$ as the spectral kernels from the $M_1$ layers. The matrices $\mathcal{A}^{(m)}$ are of rank 1 for the other $M_2$ layers and let $J^{(m)}J^{(m)T}$, where $J^{(m)}$ are matrices with orthonormal columns, be the kernels for each such $m\in \{1,\ldots,M_2\}$. Then the aggregate spectral kernel is $ K=\frac{1}{M}(M_1Z_1^T(Z_1^TZ_1)^{-1}Z_1^T + \sum_{m=1}^{M2}J^{(m)}J^{(m)T})$. Since $J^{(m)}J^{(m)T}$ is not associated with $Z$, the matrix $Z$ cannot be extracted exactly from the kernel perfectly when $M_2>0$. For small $M_2$ we can still recover a subspace close to the subspace spanned by $Z$ and the error will be governed by the Davis-Kahan theorem \cite{stewart}. However, with $M_2$ increasing eventually we will not be able to recover the subspace at all.

For OLMF, $\bar{P}=ZQ^{-1/2},\bar{\Lambda}^{(m)}=Q^{1/2}B^{(m)}Q^{1/2},m=1,\ldots,M$, is still a solution of the optimization problem (\ref{lmf}), and its uniqueness is ensured as long as at least one of the $B^{(m)}$'s in the first $M_1$ layers is of full rank.

We verify these observations on the population adjacency tensor in a simulation study, whose results are presented in Figure \ref{robust}. In the first simulation (Figure \ref{robust}(b)) we increase the number of layers uninformative of the community structure from 2 to 7 while keeping the number of informative layers fixed at 3. Spectral clustering on mean adjacency matrix continues to be able to recover the community structure perfectly. This behavior is replicated by both OLMF and co-regularized spectral clustering methods. However, aggregate of spectral kernel and module allegiance matrix approaches are not successful in recovering the correct community structure in the presence of uninformative layers. In the second simulation (Figure \ref{robust}(c)) we keep the number of informative and uninformative layers both fixed at 3 each, and vary the density of the uninformative layers while keeping the density of the informative layers fixed. As our theoretical analysis indicates, although the aggregate of spectral kernel approaches fail to recover the community structure correctly, its performance is unaffected by increasing density of the uninformative layers. Spectral clustering on mean adjacency matrix, OLMF and co-regularized spectral clustering methods continue to be able to recover the correct community structure and are unaffected by increasing density of the uninformative layers.



\vspace{10pt}

\begin{center} \textbf{\large{Appendix: Proofs}} \end{center}

\vspace{10pt}

\section*{Equivalence between problems (2.1) and (2.2)}
We denote the objective function in (2.1) as $O$. Then using properties of matrix trace and the constraint that $P^TP=I$, we have
\begin{align*}
    O & = \sum_{m=1}^{M} \tr (A^{(m)}-P\Lambda^{(m)}P^T)^T(A^{(m)}-P\Lambda^{(m)}P^T) \\
    & = \sum_{m=1}^{M} \tr (A^{(m)}A^{(m)}-2\Lambda^{(m)}P^TA^{(m)}P+P\Lambda^{(m)}P^TP\Lambda^{(m)}P^T) \\
      & = \sum_{m=1}^{M} \tr (A^{(m)}A^{(m)}-2\Lambda^{(m)}P^TA^{(m)}P+\Lambda^{(m)}\Lambda^{(m)}).
\end{align*}
Clearly, given $P$, the function $O$ is a convex function of $\Lambda^{(m)}$. Hence differentiating $O$ with respect to $\Lambda^{(m)}$ and setting it to $0$, we have
\[
\frac{\partial O}{\partial \Lambda^{(m)}} \equiv -2P^TA^{(m)}P + 2 \Lambda^{(m)} = 0,
\]
which implies that given $P$, an optimum solution of $\Lambda^{(m)}$ can be readily obtained as $\Lambda^{(m)} = P^TA^{(m)}P $. Then the OLMF problem reduces to the following optimization problem on $P$,
\begin{align*}
    & \argmin_{\{P: P^TP=I\}} \sum_{m=1}^{M} \tr (A^{(m)}A^{(m)}-2P^TA^{(m)}PP^TA^{(m)}P+P^TA^{(m)}PP^TA^{(m)}P) \\
  \equiv  &  \argmin_{\{P: P^TP=I\}} \sum_{m=1}^{M} \tr (A^{(m)}A^{(m)}-P^TA^{(m)}PP^TA^{(m)}P)  \\
  \equiv  &  \argmax_{\{P: P^TP=I\}} \sum_{m=1}^{M} \tr(P^TA^{(m)}PP^TA^{(m)}P) = \argmax_{\{P: P^TP=I\}} \sum_{m=1}^{M} \|P^TA^{(m)}P\|_F^2,
\end{align*}
which is the objective function (2.2).

\section*{Proof of Proposition 1}
\begin{proof}
Note that,
\begin{align*}
   &\|\sin \Theta(\mathbb{U}^{(m)},\mathbb{U}^{*})\|_F^2 = \frac{1}{2}\|U^{*}U^{*T}- U^{(m)}U^{(m)T}\|_F^2 \quad \quad \text{[Theorem 1.5.5 of \citep{stewart}]} \\
   & \quad = \frac{1}{2} \{ \tr(U^{*}U^{*T}U^{*}U^{*T}) + \tr( U^{(m)}U^{(m)T}U^{(m)}U^{(m)T}) -2 \, \tr(U^{*T}U^{(m)}U^{(m)T}U^{*}) \} \\
   & \quad = k- \tr(U^{*T}U^{(m)}U^{(m)T}U^{*}).
\end{align*}
Rearranging the terms we have the proposition.
\end{proof}

\section*{Proof of Lemma 1}

\begin{proof}
To prove part (i) of the lemma, note that for the case of OLMF, it is evident that $[P=Z(Z^TZ)^{-1/2},\Lambda^{(m)}=(Z^TZ)^{1/2}B^{(m)}\allowbreak (Z^TZ)^{1/2}]$ is a solution to the optimization problem on the population adjacency tensor $\mathscr{A}$. Indeed the value of the minimization objective function in (2.1) is $0$, which is its minimum possible value and $P^TP=I$. Now, by assumption, at least one of the $B^{(m)}$'s, say $B^{(m')}$, is of rank $k$. Then, without the orthogonal columns constraint on $P$ this solution is unique up to a non-singular matrix $\mu \in \mathbb{R}^{k \times k}$, since $P\mu$ and $\mu^{-1}\Lambda^{(m)}(\mu^{T})^{-1}$ for all $m$ is also a solution. However, due to the orthogonality constraint we must have, $(P\mu)^T(P\mu) =I$, which implies $\mu^T\mu=I$, i.e., $\mu$ must be an orthogonal matrix. Hence the solution is unique up to an orthogonal matrix. Moreover, since $Q^{-1/2}=(Z^TZ)^{-1/2}$ is a diagonal matrix with positive elements and hence invertible, we have $Z_{i} Q^{-1/2} =Z_{j} Q^{-1/2} \Longleftrightarrow Z_i =Z_j$.

 For co-regularized spectral clustering, note that for each $m$, maximizing  $\tr(U^{(m)T}\allowbreak\mathcal{A}^{(m)}U^{(m)})$ under the given constraints is the usual spectral clustering association cut objective function and hence is maximized by the matrix containing the $k$ eigenvectors corresponding to the $k$ algebraically largest eigenvalues of $\mathcal{A}^{(m)}$ \citep{saad11}. In this case the matrix is $\bar{U}^{(m)}=Z\mu^{(m)}$, where $\mu^{(m)}=(Z^TZ)^{-1/2}V^{(m)}$ for some orthogonal matrix $V^{(m)}$ \citep{rcy11}. Moreover by Proposition 1, it is clear that the second term has an absolute maximum value of $k$ irrespective of the first term. This maximum value is also attained with the same $\bar{U}^{(m)}$'s along with $\bar{U}^{*}=Z(Z^TZ)^{-1/2}O$, where $O$ is an orthogonal matrix. This is so because $\tr(\bar{U}^{*T}\bar{U}^{(m)}\bar{U}^{(m)T}\bar{U}^{*})=\tr (O(Z^TZ)^{-1/2}Z^TZ(Z^TZ)^{-1}Z^TZ(Z^TZ)^{-1/2}O)=k$ for all $m$. Hence, $\bar{U}^{(m)}= Z(Z^TZ)^{-1/2}V^{(m)}$ for $m=1,\ldots,M$ and $\bar{U}^{*}=Z(Z^TZ)^{-1/2}O$ are solutions up to the ambiguity of orthogonal matrices to the optimization problem. Since the matrix $Q=(Z^TZ)^{-1/2}O$ is invertible, we have $Z_{i} Q =Z_{j} Q \Longleftrightarrow Z_i =Z_j$. This gives us part (ii) of the lemma.

Next we prove part (iii) of the lemma concerning spectral clustering applied to the mean population adjacency matrix. Note that the population version is
\[
\bar{\mathcal{A}}=\frac{1}{M}\sum_{m=1}^{M}\mathcal{A}^{(m)}=\frac{1}{M}\sum_{m=1}^{M}ZB^{(m)}Z^{T} = Z \left(\frac{1}{M}\sum_{m=1}^{M}B^{(m)}\right)Z^{T}=Z\bar{B}Z^{T},
\]
with $\bar{B} \in \mathbb{R}^{k \times k}$ and is full rank as mentioned in the statement of the lemma. Then by Lemma 3.1 of \citet{rcy11}, there exists an invertible matrix $\mu \in \mathbb{R}^{k \times k}$ such that columns of $Z \mu$ are the eigenvectors of $\bar{\mathcal{A}}$, corresponding to the non-zero eigenvalues and $Z_{i} \mu =Z_{j} \mu \Longleftrightarrow Z_i =Z_j$.

Finally, for part (iv) note that if spectral clustering on aggregate spectral kernel is applied to the population adjacency tensor, each of the spectral kernels would be $Z\mu^{(m)} \mu^{(m)T} Z^{T}=Z(Z^TZ)^{-1}Z^{T}$ by the arguments in the previous paragraph. Note that the spectral kernels do not depend on $m$. Clearly, $Z(Z^TZ)^{-1/2}O$ for some orthogonal matrix $O$ is the matrix containing eigenvectors corresponding to top $k$ eigenvalues of $\bar{K}$. Denoting $(Z^TZ)^{-1/2}O$ as $Q$ we note that $Q\in \mathbb{R}^{k \times k}$ is an invertible matrix and hence $Z_{i} Q =Z_{j} Q \Longleftrightarrow Z_i =Z_j$.

\end{proof}

\section*{Proof of Theorem 1}
\begin{proof}
Our main tool to prove the result (i) will be the matrix Bernstein inequality in Theorem 5 of \citet{chung11}, which we reproduce below.

\begin{prop}
(\citet{chung11}) Let $X_1, \ldots, X_p$ be independent random $n\times n$ Hermitian matrices. Moreover, assume that $\|X_i-E(X_i)\|_2 \leq L$ for all $i$, and put $v^2=\|\sum_{i} var(X_i)\|_2$. Let $X=\sum_{i}X_i$. Then for any $a>0$,
\[
P(\|X-E(X) \|_2 > a) \leq 2n \exp \left(-\frac{a^2}{2v^2+2La/3}\right).
\]
\label{matrix_conc}
\end{prop}

Let $E^{(ij)}$ be a (deterministic) matrix with $1$ in the $(i,j)$th and $(j,i)$th position and $0$ everywhere else. Let
\[
X^{(ijm)}=(A_{ij}^{(m)}-\mathcal{A}^{(m)}_{ij})E^{(ij)}.
\]
Hence $X^{(ijm)}$ is an $n\times n$ symmetric matrix with $E(X^{(ijm)})=0$ for all $m,i,j$. Moreover, since each of $A^{(m)}_{ij}$ is an independent random variable for all $m,i,j$, the matrices $X^{(ijm)}$ are also independent. Now $\sum_m (A^{(m)}-\mathcal{A}^{(m)}) =\sum_m \sum_{ij} X^{(ijm)}$. Then following the arguments in \citet{chung11}, we have
\[
\|X^{(ijm)}\|_2 \leq 1, \quad \forall \, m,i,j,
\]
and
\begin{align*}
   v^2=\|\sum_{m} \sum_{ij} var(X^{(ijm)})\|_2 & = \|\sum_{m} \sum_{ij} E[(X^{(ijm)})^2]\|_2 \\
   & = \| \sum_{m} \sum_{ij} (\mathcal{A}^{(m)}_{ij}-(\mathcal{A}^{(m)}_{ij})^2) E^{(ii)}\|_2 \\
   & = \max_{i} \left( \sum_{m} \sum_{j} (\mathcal{A}^{(m)}_{ij} - (\mathcal{A}^{(m)}_{ij})^2)\right) \\
   & \leq \sum_m \max_{i} \sum_{j}\mathcal{A}^{(m)}_{ij}= \sum_{m} \Delta_{m} \leq M\bar{\Delta}.
\end{align*}
The third line follows since $\sum_{m} \sum_{j} (\mathcal{A}^{(m)}_{ij}-(\mathcal{A}^{(m)}_{ij})^2) E^{(ii)}$ is a diagonal matrix and hence the eigenvalues are the same as the elements.

Now we can apply the matrix concentration result in Proposition \ref{matrix_conc} to the set of independent $n \times n$ Hermitian matrices $X^{(ijm)}$. Take $a=\sqrt{4M\log (2n/ \epsilon)\bar{\Delta}}$. The assumption $M\bar{\Delta} > \frac{4}{9}\log (2n/\epsilon) $ implies that $a < 3M \bar{\Delta}$. Then applying Proposition \ref{matrix_conc} we have,
\begin{align*}
    P(\|\sum_m (A^{(m)}-\mathcal{A}^{(m)})\|_2 \geq a) & \leq 2n \exp \left(-\frac{4M \bar{\Delta}\log (2n/ \epsilon)}{2M \bar{\Delta}+2a/3}\right) \\
    & \leq 2n \exp \left(-\frac{4M\bar{\Delta}\log (2n/ \epsilon)}{4M\bar{\Delta}}\right) \\
    & \leq \epsilon.
\end{align*}
This completes the proof of part (i).

To prove part (ii) we use the following Matrix Hoeffding inequality \citep{tropp12}.
\begin{prop} (Theorem 1.3 of \citet{tropp12})
Consider a finite sequence $\{X_k\}$ of independent, random, self-adjoint matrices of common dimension $n$, and let $\{C_k\}$ be a sequence of fixed self-adjoint matrices. Assume that each of the random matrices in the sequence satisfies $E(X_k)=0$ and $X_k^2 \preceq C_k^2$ almost surely, where the symbol $\preceq$ indicates semidefinite ordering of matrices. Define $\sigma^2=\|\sum_{k}C_k^2\|_2$. Then for all $t\geq0$,
\[
P\Bigg\{\lambda_{\max} \left( \sum_k X_k\right) \geq t \Bigg \} \leq  e^{\log n- \frac{t^2}{8\sigma^2}}.
\]
\label{matrix_chernoff}
\end{prop}

To apply this bound in our case, we first note that $\{(A^{(m)}-\mathcal{A}^{(m)})^2\}$ is a sequence of independent, random and self-adjoint (Hermitian) matrices. Now, for $i \neq j$, we have
\[
E[(A^{(m)}A^{(m)})_{ij}]= E[\sum_{k} A^{(m)}_{ik}A^{(m)}_{kj}] = \sum_{k} \mathcal{A}^{(m)}_{ik}\mathcal{A}^{(m)}_{jk} = (\mathcal{A}^{(m)}\mathcal{A}^{(m)})_{ij}, \quad \forall m,
\]
and $E[(A^{(m)}A^{(m)})_{ii}]= E[\sum_{k} A^{(m)2}_{ik}]=E[\sum_{k} A^{(m)}_{ik}] =\sum_{k} \mathcal{A}^{(m)}_{ik}$ for all $m$.
Also, we have
\begin{align*}
E[\sum_m (A^{(m)}-\mathcal{A}^{(m)})^2] & = E[\sum_{m} \{A^{(m)}A^{(m)} -A^{(m)}\mathcal{A}^{(m)} -\mathcal{A}^{(m)}A^{(m)} + \mathcal{A}^{(m)}\mathcal{A}^{(m)}\}] \\
& = \sum_{m} E[A^{(m)}A^{(m)}- \mathcal{A}^{(m)}\mathcal{A}^{(m)}].
\end{align*}
Hence the matrix $E[\sum_m (A^{(m)}-\mathcal{A}^{(m)})^2]$ has 0's in all its off diagonal elements and its $n$ diagonal elements are $\sum_{m} \sum_{k} (\mathcal{A}_{ik}^{(m)}-(\mathcal{A}_{ik}^{(m)})^2)$. Define $D_m$ as the diagonal matrix whose $n$ diagonal elements are $\{\sum_{k} (\mathcal{A}_{ik}^{(m)}-(\mathcal{A}_{ik}^{(m)})^2)\}$. Then we can write
\[
\|\sum_m (A^{(m)}-\mathcal{A}^{(m)})^2\|_2 \leq \|\sum_m \{(A^{(m)}-\mathcal{A}^{(m)})^2 - D_m\} \|_2 + \|\sum_{m} D_m\|_2.
\]
Now we have
\begin{equation}
\|\sum_m D_m\|_2= \max_{i=1,\ldots ,n} \sum_{m} \sum_{k} (\mathcal{A}_{ik}^{(m)}-(\mathcal{A}_{ik}^{(m)})^2) \leq \sum_{m} \Delta_{m} =  M\bar{\Delta}.
\label{dm}
\end{equation}

Now in the notation of Proposition \ref{matrix_chernoff}, we define $X_m= (A^{(m)}-\mathcal{A}^{(m)})^2-D_m$, and $C_m^2 = \delta_{m}^2 I_{n}$, where $\delta_{m}$ is an almost sure upper bound on the largest singular value of $X_m$ and $I_{n}$ is the identity matrix of dimension $n$. Then clearly we have $X_m^2 \preceq C_m^2$ almost surely and the conditions of Proposition \ref{matrix_chernoff} is satisfied.

Next we need to find almost sure upper bound for $\| (A^{(m)}-\mathcal{A}^{(m)})^2 -D_m\|_2$ for each $m$. We use an intermediate result from the proof of Theorem 1 of \citep{chung11}, with the choice of $\epsilon=\frac{1}{n^{2}}$ and $a_m=\sqrt{4\Delta_m\log (2Mn^3)}$. For all $m$ we separately have,
\[
P(\| (A^{(m)}-\mathcal{A}^{(m)})\|_2 \geq a_m)  \leq 2n \exp \left(-\frac{a^2_m}{2\Delta_m+2a_m/3}\right).
\]
Using a union bound, at least one of the events  $\| (A^{(m)}-\mathcal{A}^{(m)})\|_2 \geq a_m$ holds with probability:
\[
2n \sum_m \exp \left(-\frac{4\Delta_m\log (2Mn^3)}{2\Delta_m+2\sqrt{4\Delta_m\log (2Mn^3)}/3}\right) \leq 2nM \frac{1}{2Mn^3} = \frac{1}{n^2}.
\]
Hence we have with probability at least $1-1/n^2$,
\[
\| (A^{(m)}-\mathcal{A}^{(m)}\|_2 \leq \sqrt{4\Delta_{m}\log(2Mn^3)}
\]
simultaneously for all $m$. Next using Borel-Cantelli Lemma, we have almost surely,
\[
\| (A^{(m)}-\mathcal{A}^{(m)})^2\|_2 = \| (A^{(m)}-\mathcal{A}^{(m)})\|_2^2 \leq 4\Delta_{m}\log(2Mn^3).
\]
Then we have,
\begin{align*}
\| ((A^{(m)}-\mathcal{A}^{(m)})^2 -D_m\|_2 & \leq 4\Delta_{m}\log(Mn^3) + \|D_m\|_2 \\
& \leq 4\Delta_{m}\log(Mn^3) + \Delta_{m} \leq 5 \Delta_{m}\log(2Mn^3).
\end{align*}
This implies,
\[ ((A^{(m)}-\mathcal{A}^{(m)})^2 -D_m)^2 \preceq (5 \Delta_{m}\log(2Mn^3))^2I_n.
\]
Finally we compute,
\[
\sigma^2=\|\sum_{m}C_m^2\|_2 = \sum_{m} \delta_m^2 =\sum_{m} 25 \Delta_{m}^2 (\log(2Mn^3))^2 = 25M\bar{\Delta}'(\log(2Mn^3))^2 .
\]

Then from Proposition \ref{matrix_chernoff} we have,
\begin{align*}
P\{ & \| \sum_m\{ (A^{(m)}-\mathcal{A}^{(m)})^2 -D_m\}\|_2 \geq  (\log n)^{(3+\epsilon)/2} \log (2M) \sqrt{M\bar{\Delta}' } \} \\
& \leq \exp\left(\log n - \frac{(\log n)^{3+\epsilon} (\log (2M))^2 M\bar{\Delta}'}{200M\bar{\Delta}'(\log(2Mn^3))^2}\right) \\
& = \exp\left(\log n - \frac{(\log n)^{3+\epsilon} (\log 2M)^2}{200(\log (2M)+3 \log n)^2}\right).
\end{align*}
If $2M>n$, the last term becomes
\begin{align*}
\exp\left(\log n - \frac{(\log n)^{3+\epsilon} (\log (2M))^2}{200(\log (2M)+3 \log n)^2}\right) \leq \exp\left(\log n - \frac{(\log n)^{3+\epsilon} (\log (2M))^2}{3200(\log (2M))^2}\right) =o(1).
\end{align*}
If $2M\leq n$, the last term becomes
\begin{align*}
\exp\left(\log n - \frac{(\log n)^{3+\epsilon} (\log (2M))^2}{200(\log (2M)+3 \log n)^2}\right) \leq \exp\left(\log n - (\log n)^{1+\epsilon} (\log (2M))^2/3200\right) =o(1).
\end{align*}

Combining the above result with Equation (\ref{dm}), we have with probability $1-o(1)$,
\[
\|\sum_{m} (A^{(m)}-\mathcal{A}^{(m)})^2\|_2 \leq (\log n)^{(3+\epsilon)/2} \log (2M) \sqrt{M\bar{\Delta}'} + M\bar{\Delta}.
\]


\end{proof}

\section*{Proof of Theorem 2}

\begin{proof}
The proof consists of three steps.
\begin{enumerate}
    \item The first step is to show that it is possible to recover the communities by maximizing the population version of the objective function.
    \item In the second step we show that for any feasible set of solutions $[\mathbb{U},U^{*}]$, the sample version of the objective function is ``close" to the population version of the objective function provided $\gamma_{m}$'s are large.
     \item Finally, in the last step we will relate the misclustering rate with the difference between $\hat{U}^{*}$ and $\bar{U}^{*}$, and then relate this difference with the difference between the maximized sample and the population versions of the objective function.
\end{enumerate}

The result of Lemma 1 shows that  $\bar{U}^{(m)}=Z\mu^{(m)}, \, m=1,\ldots,M, \, \bar{U}^{*}=Z(Z^TZ)^{-1/2}O$ is the solution up to the ambiguity of (several different) orthogonal matrices obtained by optimizing the population version of the objective function $F(\mathscr{A},\mathbb{U},U^{*})$. We call the tensor containing the layer-wise low rank matrices, $\bar{U}^{(m)}$, as $\mathbb{\bar{U}}$. Note that $\bar{U}^{(m)}=\bar{U}^{*}V^{(m)}$, for some orthogonal matrix $V^{(m)}$. Lemma 1 further shows that the true community assignments $Z$ can be recovered by applying k-means algorithm to the columns of $\bar{U}^{*}$.

Let $[\mathbb{U},U^{*}]$ be a feasible set of solutions. Then we have with probability at least $1-\epsilon/2$,
\begin{align*}
    |\sum_{m}\tr(U^{*T}(\mathcal{A}^{(m)}-A^{(m)})U^{*})| & \leq k \|\sum_{m}U^{*T}(\mathcal{A}^{(m)}-A^{(m)})U^{*}\|_2 \\
    & \leq k \|U^{*}\|_2^2\|\sum_{m}(\mathcal{A}^{(m)}-A^{(m)})\|_2  \\
    & \leq k \sqrt{4M \bar{\Delta}\log (4n/\epsilon)},
\end{align*}
where the first inequality is true since $\sum_{m} U^{*T}(\mathcal{A}^{(m)}-A^{(m)})U^{*}$ is a $k \times k$ matrix, the second line follows since $\|AB\|_2 \leq \|A\|_2 \|B\|_2$ for any two matrices $A$ and $B$, while the third inequality follows from Theorem 1.

We define two square symmetric $k \times k$ matrices, $\bar{S}^{(m)}= \bar{U}^{*T}\mathcal{A}^{(m)}\bar{U}^{*}$ and $\hat{S}^{(m)}= \hat{U}^{*T}\mathcal{A}^{(m)}\hat{U}^{*}$. Since $\bar{U}^{*}V^{(m)}$ is the matrix of eigenvectors corresponding to the non-zero eigenvalues of $\mathcal{A}^{(m)}$, we also have the eigenvalue decomposition, $\mathcal{A}^{(m)}=\bar{U}^{*}\bar{S}^{(m)}\bar{U}^{*T}$. We define a new quantity $\mathcal{A}_{1}^{(m)}=\hat{U}^{*}\hat{S}^{(m)}\hat{U}^{*T}=\hat{U}^{*}\hat{U}^{*T}\mathcal{A}^{(m)}\hat{U}^{*}\hat{U}^{*T}$.  Then $\hat{U}^{*}$ is an invariant subspace of $\mathcal{A}^{(m)}_1$ \citep{pc16}. A couple of lines of algebra show that (see \citet{pc16} for a proof)
\begin{equation}
\|\mathcal{A}^{(m)}-\mathcal{A}_{1}^{(m)}\|_F^2=\|\bar{S}^{(m)}\|_F^2-\|\hat{S}^{(m)}\|_F^2.
\label{frompc16}
\end{equation}

For a $k \times k$ matrix $B$, let $\lambda_1 \geq \lambda_2 \geq \cdots \geq \lambda_k$ be the eigenvalues of $B$ sorted in decreasing order. Since the eigenvalues of $\bar{S}^{(m)}$ are the non-zero eigenvalues of $\mathcal{A}^{(m)}$, we also have the following eigenvalue interlacing property (Theorem 2.1 of \citet{haemers95}),
\begin{equation}
\lambda_i(\bar{S}^{(m)}) = \lambda_i(\bar{U}^{*T}\mathcal{A}^{(m)}\bar{U}^{*}) \geq \lambda_i(\hat{U}^{*T}\mathcal{A}^{(m)}\hat{U}^{*}) = \lambda_i(\hat{S}^{(m)}),
\label{interlacing}
\end{equation}
for all $1 \leq i \leq k$.
Then we have the following bound on the Frobenius norm of difference between $\mathcal{A}^{(m)}$ and $\mathcal{A}_{1}^{(m)}$ in terms of the traces of $\bar{S}^{(m)}$ and $\hat{S}^{(m)}$:
\begin{align}
    \|\mathcal{A}^{(m)}-\mathcal{A}_{1}^{(m)}\|_F^2 & =\sum_{i=1}^{k}\lambda^2_i(\bar{S}^{(m)})-\sum_{i=1}^{k}\lambda^2_i(\hat{S}^{(m)}) \quad \quad \text{[Equation (\ref{frompc16})]}\nonumber\\
    & \leq \sum_{i=1}^{k} |\lambda_i(\bar{S}^{(m)})-\lambda_i(\hat{S}^{(m)}) | |\lambda_i(\bar{S}^{(m)})+\lambda_i(\hat{S}^{(m)})| \nonumber\\
    & \leq \sum_{i=1}^{k}|\lambda_i(\bar{S}^{(m)})-\lambda_i(\hat{S}^{(m)}) | \cdot 2 |\lambda_i(\bar{S}^{(m)})| \quad \quad \text{[Property (\ref{interlacing})]} \nonumber\\
    & \leq 2 |\lambda_1(\bar{S}^{(m)})| \sum_{i=1}^{k} (\lambda_i(\bar{S}^{(m)})-\lambda_i(\hat{S}^{(m)})) \quad \quad \text{[Property (\ref{interlacing})]} \nonumber\\
    & \leq 2 \Delta_m ( \tr (\bar{S}^{(m)}) - \tr(\hat{S}^{(m)})) \quad \quad \text{[Since } \lambda_1(\bar{S}^{(m)}) \leq \Delta_m \text{]}.
    \label{trace}
\end{align}

Finally we use this result to prove the following bound which then leads to a bound on misclustering rate:
\begin{align*}
   \sum_{m} & \frac{(\lambda^{(m)})^2}{2 \Delta_m} \|\hat{U}^{*}-\bar{U}^{*}O\|_F^2  \\
      & \leq \sum_{m} \frac{1}{2 \Delta_m}\|\mathcal{A}^{(m)}-\mathcal{A}_{1}^{(m)}\|_F^2 \quad \text{[Davis-Kahan Theorem in \citep{stewart}]}  \\
   & \leq \sum_{m} \tr(\bar{U}^{*T}\mathcal{A}^{(m)}\bar{U}^{*}-\hat{U}^{*T}\mathcal{A}^{(m)}\hat{U}^{*}) \quad \text{[Equation (\ref{trace})]} \\
   & \leq \sum_{m} \{\tr(\bar{U}^{*T}\mathcal{A}^{(m)}\bar{U}^{*}-\hat{U}^{*T}\mathcal{A}^{(m)}\hat{U}^{*})  \\
   & \quad + \tr ( \hat{U}^{(m)T}A^{(m)}\hat{U}^{(m)} -\bar{U}^{(m)T}A^{(m)}\bar{U}^{(m)} ) + \gamma_{m} (k-\frac{1}{2} \|\hat{U}^{(m)}\hat{U}^{(m)T}-\hat{U}^{*}\hat{U}^{*T}\|_F^2-k) \} \\
  &  = \sum_{m} \{\tr(\bar{U}^{*T}\mathcal{A}^{(m)}\bar{U}^{*}-\bar{U}^{*T}A^{(m)}\bar{U}^{*})+ \tr ( \hat{U}^{*T}A^{(m)}\hat{U}^{*} - \hat{U}^{*T}\mathcal{A}^{(m)}\hat{U}^{*})\} \\
    & \quad + \sum_{m} \{\tr ( \hat{U}^{(m)T}A^{(m)}\hat{U}^{(m)} -\hat{U}^{*T}A^{(m)}\hat{U}^{*} ) -\gamma_{m}\frac{1}{2} \|\hat{U}^{(m)}\hat{U}^{(m)T}-\hat{U}^{*}\hat{U}^{*T}\|_F^2\} \\
   & \leq 2k \sqrt{4M \bar{\Delta}\log (4n/\epsilon)}+ \sum_{m} \{\sum_{j} |\lambda_{j}(\hat{U}^{(m)T}A^{(m)}\hat{U}^{(m)}) -\lambda_j(\hat{U}^{*T}A^{(m)}\hat{U}^{*})| \\
   &  \quad -\gamma_{m} \|\sin \Theta (\hat{U}^{(m)},\hat{U}^{*})\|_F^2\} \\
   & \leq 2k \sqrt{4M \bar{\Delta}\log (4n/\epsilon)}+ \sum_{m}(\|A^{(m)}\|_2 \|\sin \Theta (\hat{U}^{(m)},\hat{U}^{*})\|_{\Sigma}-\gamma_{m} \|\sin \Theta (\hat{U}^{(m)},\hat{U}^{*})\|_F^2),
\end{align*}
with probability at least $1-\epsilon$. The third inequality follows from the fact that $F(\mathbb{A},\mathbb{\hat{U}},\hat{U}^{*}) \geq F(\mathbb{A},\mathbb{\bar{U}},\bar{U}^{*})$ and Proposition 1. The first term in the fourth equality has used the fact: since $V^{(m)}$'s are orthogonal matrices, $\tr(V^{(m)T}\bar{U}^{*T}A^{(m)}\bar{U}^{*}V^{(m)})=\tr(\bar{U}^{*T}A^{(m)}\bar{U}^{*})$. The last line follows from Theorem 2.1 of \citet{knyazev10} which states that $\sum_{j} |\lambda_{j}(\hat{U}^{(m)T}A^{(m)}\hat{U}^{(m)}) -\lambda_j(\hat{U}^{*T}A^{(m)}\hat{U}^{*})|  \leq \sum_j \|A^{(m)}\|_2 \allowbreak \sin   \theta_j (\hat{U}^{(m)},\hat{U}^{*})$ and Proposition 1. Since $\sin \Theta$ is a diagonal matrix with non-negative elements, we represent $\sum_j \sin \theta_j$ as $\|\sin \Theta\|_{\Sigma}$.

Now, if $ \max_{j,m}|\sin \theta^{(m)}_j| \leq \frac{ \sqrt{4 \bar{\Delta}\log (4n/\epsilon)}}{\sqrt{M}\| A^{(m)}\|_2}$, then we have, $ \sum_{m}(\|A^{(m)}\|_2 \|\sin \Theta (\hat{U}^{(m)},\hat{U}^{*})\|_{\Sigma} = \sum_m \sum_j \|A^{(m)}\|_2 |\sin \theta^{(m)}_j| \leq k \sqrt{4M \bar{\Delta}\log (4n/\epsilon)}$ and consequently, the last line is upper bounded by $3k \sqrt{4M \bar{\Delta}\log (2N/\epsilon)}$. On the other hand, if $ \min_{j,m}|\sin \theta^{(m)}_j| > \frac{ \sqrt{4 \bar{\Delta}\log (4n/\epsilon)}}{\sqrt{M}\| A^{(m)}\|_2}$, then the last two terms together go to 0, if $\frac{\gamma_{m}}{\|A^{(m)}\|_2}|\sin \theta^{(m)}_j| >1$ for all $j$, which implies
\[
\gamma_{m}> \frac{\|A^{(m)}\|_2}{\min_{j,m} |\sin \theta^{(m)}_j|} = \frac{\sqrt{M} \|A^{(m)}\|_2^2}{\sqrt{4 \bar{\Delta}\log (4n/\epsilon)}}.
\]

Together the above two cases define an exhaustive set of all possible values $\sin \theta^{(m)}_j$ can take. Hence if we choose $\gamma_{m}$ to be large enough such that $\gamma_{m} >  \frac{\sqrt{M} \|A^{(m)}\|_2^2}{\sqrt{4 \bar{\Delta}\log (4n/\epsilon)}}$, then we have with probability at least $(1-\epsilon)$,
\[
    \|\hat{U}^{*}-\bar{U}^{*}O\|_F^2  \leq \frac{6k \sqrt{4M \bar{\Delta}\log (4n/\epsilon)}}{\sum_{m} \frac{(\lambda^{(m)})^2}{\Delta_m}}.
\]

The bound on misclustering rate follows:
\[
r_{coreg} \leq \frac{8n_{\max}}{n}\|\hat{U}-\bar{U}^{*}O\|_F^2 \leq \frac{96n_{\max}k}{n\frac{1}{M}\sum_{m} \frac{(\lambda^{(m)})^2}{\Delta_m}} \sqrt{\frac{\bar{\Delta}\log (4n/\epsilon)}{M}},
\]
with probability at least $1-\epsilon$.
\end{proof}

\section*{Proof of Lemma 2}

We need to show that we can replace $\bar{\Delta}$ in the denominator with $\frac{1}{2M}\sum_{m}\|A^{(m)}\|_2$ and the resulting bound on $\gamma_m$ will hold with high probability.
We have,
\[
\bar{\Delta} = \frac{1}{M}\sum_m\Delta_m \geq \frac{1}{M}\sum_m\|\mathcal{A}^{(m)}\|_2 \geq \|\frac{1}{M}\sum_m\mathcal{A}^{(m)}\|_2.
\]
Now we further have,
\begin{align*}
    P(&\|\frac{1}{M}\sum_mA^{(m)}\|_2 > 2\bar{\Delta}) \\
    & =  P(\|\frac{1}{M}\sum_m A^{(m)}\|_2 -\|\frac{1}{M}\sum_m\mathcal{A}^{(m)}\|_2 > 2\bar{\Delta}-\|\frac{1}{M}\sum_m\mathcal{A}^{(m)}\|_2) \\
    & \leq P(\|\frac{1}{M}\sum_m A^{(m)}\|_2 -\|\frac{1}{M}\sum_m\mathcal{A}^{(m)}\|_2 > \bar{\Delta}) \\
        & \leq P(\|\frac{1}{M}\sum_m(A^{(m)}-\mathcal{A}^{(m)})\|_2 > \bar{\Delta}) \\
   &  \leq 2n\exp\left(-\frac{M^2\bar{\Delta}^2}{M\bar{\Delta} + 2M\bar{\Delta}/3}\right) \\
   & \leq 2n\exp\left(-\frac{3}{5}M\bar{\Delta}\right) \leq 2n\frac{\delta}{2n} = \delta.
\end{align*}
Hence with probability at least $1-\delta$, we have $\|\frac{1}{2M}\sum_m A^{(m)}\|_2 \leq \bar{\Delta} $ and hence we can replace $\bar{\Delta}$ with $\|\frac{1}{2M}\sum_m A^{(m)}\|_2 \leq \bar{\Delta} $ in the denominator of the expression for condition required on $\gamma_m$.

\section*{Proof of Lemma 3}
\begin{proof}
We note that for the four parameters MLSBM,
\[
\mathcal{A}^{(m)}=Z(Z^TZ)^{-1/2}(Z^TZ)^{1/2}B^{(m)} (Z^TZ)^{1/2}(Z^TZ)^{-1/2}Z^T=HS^{(m)}H^T,
\]
where $H=Z(Z^TZ)^{-1/2}$ and $S^{(m)}=(Z^TZ)^{1/2}B^{(m)} (Z^TZ)^{1/2}$. Clearly \\$\mathcal{A}^{(m)}H=HS^{(m)}$, and hence columns of $H$ span a $k$ dimensional invariant subspace of $\mathcal{A}^{(m)}$. Moreover since $\text{rank}(\mathcal{A}^{(m)})=\text{rank}(S^{(m)})$, all non-zero eigenvalues of $\mathcal{A}^{(m)}$ are also eigenvalues of $S^{(m)}$. This implies the smallest non-zero eigenvalue of $\mathcal{A}^{(m)}$ is also the smallest eigenvalue of $S^{(m)}$. To determine the smallest eigenvalue we proceed as in \citet{rcy11}. Note that we have $S^{(m)}=\sqrt{s}I_kB^{(m)}\sqrt{s}I_k=sB^{(m)}$, and $B^{(m)}$ can be written as $B^{(m)}=(p^{(m)}-q^{(m)})I_k+q^{(m)}1_k1_k^T$. Then $1_k$ is an eigenvector of $S^{(m)}$ since $sB^{(m)}1_k=(s(p^{(m)}-q^{(m)})+sq^{(m)}k)1_k=(s(p^{(m)}-q^{(m)})+nq^{(m)})1_k$. Let $u$ be another eigenvector of $sB^{(m)}$. Then $\|u\|_2=1$ and $u^T1_k=0$. Hence we have $sB^{(m)}u=s(p^{(m)}-q^{(m)})u$. This implies all the remaining eigenvalues of $sB^{(m)}$ are $s(p^{(m)}-q^{(m)})$. Since $nq^{(m)}>0$, we conclude the smallest eigenvalue of $sB^{(m)}$ is $s(p^{(m)}-q^{(m)})$. This is also the smallest non-zero eigenvalue of $\mathcal{A}^{(m)}$.
\end{proof}

\section*{Proof of Theorem 3}

\begin{proof}
Similar to the proof of Theorem 2, the proof for this theorem also consists of three steps. The first step was addressed in Lemma 1, where it was shown that true community labels can be recovered from the solution $\bar{P}$ of the objective function applied to the population adjacency tensor. Next we show the second step. For any feasible solution of $P$ we have,
\begin{align*}
     | & F(\mathscr{A},P)  - F(\mathbb{A},P) |  = |\sum_{m} \{\|P^{T}\mathcal{A}^{(m)}P\|^2_F-\|P^{T}A^{(m)}P\|^2_F \} |\\
     & = \sum_{m}\{ (\|P^{T}A^{(m)}P\|_F-\|P^{T}\mathcal{A}^{(m)}P\|_F )^2 \\
     & \quad + | (\|P^{T}A^{(m)}P\|_F-\|P^{T}\mathcal{A}^{(m)}P\|_F)\cdot 2 \|P^{T}\mathcal{A}^{(m)}P\|_F | \} \\
     & = \sum_{m} \{(x^{(m)}-y^{(m)})^2 + |2y^{(m)}(x^{(m)}-y^{(m)})|\} \\
     & = \sum_{m} \{(x^{(m)}-y^{(m)})^2 + 2|y^{(m)}| \, |(x^{(m)}-y^{(m)})|\}
\end{align*}
where $x^{(m)}=\|P^{T}A^{(m)}P\|_F$ and $y^{(m)}=\|P^{T}\mathcal{A}^{(m)}P\|_F$.

First, for the $2|y^{(m)}|$ term we have,
\begin{align*}
|y^{(m)}| & = \|P^{T}\mathcal{A}^{(m)}P\|_F \\
& \leq \sqrt{k}\|P^{T}\mathcal{A}^{(m)}P\|_2 \quad \text{[equivalence of norm since $P^{T}\mathcal{A}^{(m)}P$ is $k \times k$]} \\
& \leq \sqrt{k} \|P\|^2_2 \|\mathcal{A}^{(m)}\|_2 \quad \text{[property of spectral norm, $\|AB\|_2 \leq \|A\|_2\|B\|_2$]}\\
& \leq \sqrt{k}\Delta_{m} \quad \text{[since $\|\mathcal{A}^{(m)}\| \leq \Delta_{m}$]}.
\end{align*}
Now, since $\|A\|_F -\|B\|_F \leq \|A-B\|_F$, we have
\[
\sum_{m} 2|y^{(m)}||(x^{(m)}-y^{(m)})| \leq \sqrt{k} \sum_{m}\Delta_{m}  \|P^{T}(A^{(m)}-\mathcal{A}^{(m)})P\|_F.
\]

Then using Cauchy-Schwartz inequality we have the following result,
\begin{align*}
  \sqrt{k} \sum_{m} & \Delta_{m} \|P^{T}(A^{(m)}-\mathcal{A}^{(m)})P\|_F  \\
   & \leq  \sqrt{k} \sum_{m} \Delta_{m}  \sqrt{\tr(P^{T}(A^{(m)}-\mathcal{A}^{(m)})PP^{T}(A^{(m)}-\mathcal{A}^{(m)})P)} \\
    & \leq \sqrt{k} \sqrt{\sum_m \Delta_{m}^2 }\sqrt{\tr(\sum_{m}P^T (A^{(m)}-\mathcal{A}^{(m)})PP^{T}(A^{(m)}-\mathcal{A}^{(m)}))P)}  \\
    & \leq \sqrt{k} \sqrt{M\bar{\Delta}'} \sqrt{\|PP^T\|_2 \tr(\sum_{m} (A^{(m)}-\mathcal{A}^{(m)})PP^{T}(A^{(m)}-\mathcal{A}^{(m)})) } \\
    & = \sqrt{k} \sqrt{M\bar{\Delta}'}  \sqrt{\tr(P^T\sum_{m} (A^{(m)}-\mathcal{A}^{(m)})(A^{(m)}-\mathcal{A}^{(m)})P)} \\
    & \leq \sqrt{k} \sqrt{M\bar{\Delta}'}  \sqrt{  k\|P\|_2^2 \|\sum_{m}(A^{(m)}-\mathcal{A}^{(m)})(A^{(m)}-\mathcal{A}^{(m)})\|_2 } \\
    & \leq \sqrt{k} \sqrt{M\bar{\Delta}'}  \sqrt{k(M\bar{\Delta} + \sqrt{M \bar{\Delta}'} \log 2M(\log n)^{2+\epsilon}} \\
    & \leq kM^{3/4} (\log 2M)^{1/2}(\log n)^{1+\epsilon/2} (\bar{\Delta}')^{3/4} + kM (\bar{\Delta}')^{1/2}(\bar{\Delta})^{1/2},
\end{align*}
with probability at least $1-o(1)$.
In the above result, the inequality in line 3 is due to Cauchy-Schwartz inequality and line 4 follows from the inequality on trace of product of a positive semi-definite matrix ($(A^{(m)}-\mathcal{A}^{(m)})PP^{T}(A^{(m)}-\mathcal{A}^{(m)})$) with a Hermitian matrix ($PP^T$) due to \citet{wang1986trace} (See also \citet{fang1994inequalities}). The inequality in line 6 follows from the relations $\tr(XY) \leq k\|XY\|_2 \leq k\|X\|_2 \|Y\|_2$. Finally the inequality in line 7 follows from Theorem 1 part (ii).

Similarly, we can derive
\begin{align*}
\sum_{m} (x^{(m)}-y^{(m)})^2 & \leq \sum_{m} \|P^{T}(A^{(m)}-\mathcal{A}^{(m)})P\|_F^2 \\
& = \sum_{m} \tr((P^{T}(A^{(m)}-\mathcal{A}^{(m)})PP^{T}(A^{(m)}-\mathcal{A}^{(m)})P)) \\
& \leq k(M\bar{\Delta} + \sqrt{M \bar{\Delta}'} \log 2M(\log n)^{2+\epsilon}),
\end{align*}
with probability at least $1-o(1)$.

Finally, since $(\log 2M)^{1/2} = o(M^{1/4})$, and $\bar{\Delta} < \bar{\Delta}'$, combining the results together we have with probability at least $1-o(1)$,
\[
| F(\mathscr{A},P)  - F(\mathbb{A},P) | \leq 3kM (\bar{\Delta}')^{1/2}(\bar{\Delta})^{1/2} + 3kM^{3/4} (\log 2M)^{1/2}(\log n)^{2+\epsilon} (\bar{\Delta}')^{3/4}.
\]

Let $\hat{P}$ be the solution of the optimization problem in OLMF, i.e., $\hat{P}$ maximizes $F(\mathbb{A},P)$. Further let $\bar{P}$ maximizes the population version of the objective function $F(\mathscr{A},P)$. Then $F(\mathbb{A},\hat{P}) \geq F(\mathbb{A},\bar{P}) $, and $F(\mathscr{A},\bar{P}) \geq F(\mathscr{A},\hat{P}) $. Consequently, we have with probability at least  $1-o(1)$,
\begin{align*}
F(\mathscr{A},\bar{P}) - F(\mathscr{A},\hat{P}) & \leq F(\mathscr{A},\bar{P}) - F(\mathscr{A},\hat{P}) + F(\mathbb{A},\hat{P}) - F(\mathbb{A},\bar{P}) \\
& \leq |F(\mathscr{A},\bar{P}) - F(\mathbb{A},\bar{P})| + |F(\mathscr{A},\hat{P}) -F(\mathbb{A},\hat{P})| \\
& \leq 6 kM^{3/4} \bar{\Delta}^{'1/2}( M^{1/4}\bar{\Delta}^{1/2} + (\log 2M)^{1/2}(\log n)^{2+\epsilon} \bar{\Delta}'^{1/4}).
\end{align*}

Now define $\bar{\Lambda}^{(m)}=\bar{P}^{T}\mathcal{A}^{(m)}\bar{P}$ and $\Lambda_1^{(m)}=\hat{P}^{T}\mathcal{A}^{(m)}\hat{P}$. Note that since $\bar{P}$ is an invariant subspace of $\mathcal{A}^{(m)}$, we have $\mathcal{A}^{(m)}=\bar{P}^{T}\bar{\Lambda}^{(m)}\bar{P}$. We define $\mathcal{A}_1^{(m)}=\hat{P}\Lambda_1^{(m)}\hat{P}^{T}=\hat{P}\hat{P}^{T}\mathcal{A}^{(m)}\hat{P}\hat{P}^{T}$. Then $\hat{P}$ is an invariant subspace of $\mathcal{A}_1^{(m)}$. Further, we have for all $m$,
\[
\|\bar{P}^{T}\mathcal{A}^{(m)}\bar{P}\|_F^2-\|\hat{P}^{T}\mathcal{A}^{(m)}\hat{P}\|_F^2 = \|\mathcal{A}^{(m)}-\mathcal{A}_1^{(m)}\|_F^2. \quad \text{[\citet{pc16}]}
\]
This result along with (3.2) imply,
\begin{align*}
    F & (\mathscr{A},\bar{P}) - F(\mathscr{A},\hat{P})  = \sum_{m} \|\bar{P}^{T}\mathcal{A}^{(m)}\bar{P}\|_F^2-\|\hat{P}^{T}\mathcal{A}^{(m)}\hat{P}\|_F^2 \\
    & = \sum_{m} \|\mathcal{A}^{(m)}-\hat{P}\hat{P}^{T}\mathcal{A}^{(m)}\hat{P}\hat{P}^{T}\|_F^2
     \geq \sum_{m} (\lambda^{(m)})^2\|\hat{P}-\bar{P}O\|_F^2
     \geq \frac{nr_{LMF}}{8n_{\max}} \sum_{m} (\lambda^{(m)})^2.
\end{align*}
Hence we have with probability at least $1-o(1)$
\[
r_{LMF} \leq \frac{48n_{\max}k\bar{\Delta}^{'1/2}(\bar{\Delta}^{1/2} + \bar{\Delta}'^{1/4}(\log 2M)^{1/2}(\log n)^{2+\epsilon}/M^{1/4} )}{\frac{1}{M}\sum_{m} (\lambda^{(m)})^2 n}.
\]
\end{proof}

\section*{Proof of Theorem 4}
\begin{proof}

We use the bound on the quantity $\|\bar{A}-\bar{\mathcal{A}}\|_2$ obtained in Theorem 1 part (i), Lemma 5.1 in \citep{lei14} and the Davis-Kahan Theorem \citep{dk70} to obtain the following bound:
\begin{align*}
    \|\hat{U}-\bar{U}O\|_F \leq \frac{2 \sqrt{2}\sqrt{k} \|\bar{A}-\bar{\mathcal{A}}\|_2}{\lambda^{\bar{\mathcal{A}}}}  & \leq \frac{2\sqrt{2}\sqrt{k}}{\lambda^{\bar{\mathcal{A}}}}  \sqrt{\frac{4\bar{\Delta}\log (2n/ \epsilon)}{M}}
     \\
     & = \frac{4\sqrt{2}}{\lambda^{\bar{\mathcal{A}}}}\sqrt{\frac{k \bar{\Delta}\log (2n/ \epsilon)}{M}},
\end{align*}
with probability at least $1-\epsilon$ for any $\epsilon>0$. Hence using (3.2) the misclustering rate is bounded as
\[
r_{av} \leq \frac{8n_{\max}}{n}\|\hat{U}-\bar{U}O\|_F^2 \leq \frac{256 n_{\max}k \bar{\Delta}\log (2n/ \epsilon) }{(\lambda^{\bar{\mathcal{A}}})^2nM},
\]
with probability at least $1-\epsilon$.
\end{proof}

\section*{Proof of Lemma 4}
\begin{proof}
From the arguments in the proof of Lemma 3 we have,
\[
\bar{B}=\frac{1}{M}\sum_{m} B^{(m)}=\frac{1}{M} \sum_{m}\{(p^{(m)}-q^{(m)})I_k + q^{(m)}1_k1_k^T\}.
\]
Hence $1_k$ is an eigenvector of $\bar{B}$ corresponding to the largest eigenvalue $s\frac{1}{M}\sum_{m} \{(p^{(m)}-q^{(m)})+nq^{(m)}\}$. All other eigenvectors correspond to the eigenvalue $s\frac{1}{M}\sum_{m} (p^{(m)}-q^{(m)})$. Hence $ \lambda^{\bar{\mathcal{A}}}=s\frac{1}{M}\sum_{m} (p^{(m)}-q^{(m)})$.
\end{proof}

\bibliography{msp}

\end{document}